\documentclass[twoside]{article}

\usepackage{amsmath,amsfonts,bm}

\def\eqref#1{equation~\ref{#1}}

\def\1{\bm{1}}

\def\ra{{\textnormal{a}}}

\DeclareMathAlphabet{\mathsfit}{\encodingdefault}{\sfdefault}{m}{sl}
\SetMathAlphabet{\mathsfit}{bold}{\encodingdefault}{\sfdefault}{bx}{n}

\def\sS{{\mathbb{S}}}

\newcommand{\E}{\mathbb{E}}

\newcommand{\R}{\mathbb{R}}

\newcommand{\KL}{D_{\mathrm{KL}}}

\DeclareMathOperator*{\argmax}{arg\,max}
\DeclareMathOperator*{\argmin}{arg\,min}

\newcommand{\latentsymbolspace}{\sS'}

\usepackage[accepted]{aistats2023}
\usepackage{natbib}
\usepackage[utf8]{inputenc} 
\usepackage[T1]{fontenc}    
\usepackage{url}            
\usepackage{booktabs}       
\usepackage{amsfonts}       
\usepackage{amssymb}
\usepackage{amsthm}
\usepackage{nicefrac}       
\usepackage{microtype}      
\usepackage{lipsum}		
\usepackage{graphicx}
\usepackage{caption}
\usepackage{subcaption}
\usepackage{bbm}

\usepackage{xcolor}
\definecolor{linkcolor}{RGB}{74, 102, 146}
\usepackage[colorlinks=true,allcolors=linkcolor,pageanchor=true,plainpages=false,pdfpagelabels,bookmarks,bookmarksnumbered]{hyperref}

\usepackage[ruled,vlined]{algorithm2e}
\include{pythonlisting}

\usepackage{doi}
\usepackage[short]{optidef}
\usepackage{cleveref}
\usepackage{thmtools,thm-restate}

\usepackage{xspace}
\newcommand*{\eg}{{\it e.g.}\@\xspace}
\newcommand*{\ie}{{\it i.e.}\@\xspace}

\newcommand{\T}{\mathcal{T}}

\renewcommand{\ra}[1]{\renewcommand{\arraystretch}{#1}}

\newcommand{\methodname}{Latent VD-SDE\xspace}

\begin{document}

%

%

\twocolumn[

\aistatstitle{Latent Discretization for Continuous-time Sequence Compression}

\aistatsauthor{Ricky T. Q. Chen, Matthew Le, Matthew Muckley, Maximilian Nickel, Karen Ullrich}

\aistatsaddress{ \texttt{\{rtqichen,mattle,mmuckley,maxn,karenu\}@meta.com} \\ Meta AI } ]

\begin{abstract}
Neural compression offers a domain-agnostic approach to creating codecs for lossy or lossless compression via deep generative models. 
For sequence compression, however, most deep sequence models have costs that scale with the sequence length rather than the sequence complexity. 
In this work, we instead treat data sequences as observations from an underlying continuous-time process and learn how to efficiently discretize while retaining information about the full sequence.
As a consequence of decoupling sequential information from its temporal discretization, our approach allows for greater compression rates and smaller computational complexity.
Moreover, the continuous-time approach naturally allows us to decode at different time intervals.
We empirically verify our approach on multiple domains involving compression of video and motion capture sequences, showing that our approaches can automatically achieve reductions in bit rates by learning how to discretize.


\end{abstract}

\section{Introduction}

Sequential data is prevalent in the world as data collected from very different modalities all commonly have a temporal aspect and potentially complex temporal dependencies. As such, sequence modeling has become a central research topic in machine learning, specifically in areas such as healthcare, neuroscience, natural language processing, graphics and virtual reality, among many others. Furthermore, it has growing importance for real world applications. For instance, the global IP traffic tripled from 2016 to 2021 largely due to increased video streaming making up a majority of that traffic~\citep{barnett2018cisco,barnett2021cisco}. Thus concurrently, facilitating the growing needs to store and communicate sequences is a new frontier of machine learning research in the area of neural compression, \ie efficiently storing data using deep sequence models.

In this work, we center the discussion around data sequences that are inherently observed from an underlying continuous signal.
Often, these sequences are discretized in time according to a domain-specific or manually selected frequency.
Instead, we hypothesize that if we can recover the underlying continuous signal, then we can also learn how to efficiently discretize this signal without the need for domain knowledge or manual tuning. 

Specifically, we use a latent stochastic differential equation (SDE) model~\citep{oksendal2003stochastic,sarkka2013bayesian} to infer a latent continuous-time process that represents the data sequence.
A discretization model based on point processes~\citep{daley2003introduction} then simultaneously learns how to discretize the continuous-time representation depending on its complexity.
Furthermore, the SDE approach allows us to place a highly interpretable prior on the latent trajectories that separates states into global and local representations, which can be further used to systematically reduce bit rate in the lossy compression regime.
Our method can be understood as a deep generative model of sequential data that compactly compresses data sequences. In contrast to other methods that undergo extensive manual hyper-parameter tuning to achieve a desired rate-distortion trade-off, our method moves along its rate-distortion curve seamlessly and automatically without the need of training multiple models. 
Our contribution is to think about compression not starting from a fixed discretization (i.e. fixed frames per second) but rather to discuss how the original or underlying continuous signal can be best discretized; this allows us to compress and decompress sequences without relying on an arbitrary fixed frames per second.

\section{Background \& Preliminaries}

\subsection{Lossy Compression with Sequential Latent Variable Models}

The goal of lossy compression is to encode data points $x$ from a distribution $p_d(x)$ over discrete space $\mathcal{X}$ as bit strings $m = \texttt{enc}(x) \in \{0,1\}^*$ of shortest possible length $\ell(m)$ such that $x$ when decoded  $x^\prime=\texttt{dec}(\texttt{enc}(x))$ experiences only a small loss of information as quantified by any distortion measure $d(x,x^\prime)$, e.g. the mean squared error.
The tuple $(\texttt{enc},\texttt{dec})$ is referred to as a codec. 
More formally,  \citet{shannon1948mathematical} describes the lossy compression problem through a rate-distortion (RD) theory,
that characterizes the optimal bit-rate by the following minimization problem  
\begin{align}
\small
    \ell(m) \geq \underset{q(z|x)}{\inf} \text{  } I(z;x)  &\text{ subject to } d(x, x^\prime) \leq \delta,
\end{align}
where $q(z|x)$ is any arbitrary distribution referred to as test channel,
$I(.;.)$ is the mutual information (MI) and $\delta$ the maximum level of tolerable distortion.   
Solving this optimization problem directly is infeasible as the MI is notoriously hard to compute \citep{alemi2017fixing,barber2003information,alemi2016deep,agakov2005variational}. Typically, the MI is bounded as follows 
\begin{align}\small
    I(z;x) \leq \E_{x\sim p_d(x)}\left[\KL(q(z|x)||p(z))\right],
\end{align}
where $p(z)$ is an approximation to the marginal posterior $q(z)= \E_{x\sim p_d(x)}\left[q(z|x)\right]$. 
Practical realizations of this rate have recently been discussed \citep{li2018strong,flamich2020compressing,flamich2022fast,theis2021algorithms} and are currently referred to as relative entropy coding (REC). 
However, due to computational constraints of these methods especially for high dimensional encodings, this group of algorithms is of practical interest only for low dimensional variables.

In contrast to REC algorithms, traditional lossy compression algorithms first map $x$ to a latent $z$, and subsequently perform lossless compression of $z$ using entropy coders. 
This method results in a sub-optimal rate $H(q(z|x), p(z))\geq \KL(q(z|x)||p(z))$, where $H(q(z|x), p(z)) \triangleq \E_{q(z|x)}[-\log p(z)]$ is the cross entropy. See Appendix \ref{app:lossless} for more details on lossless compression.

Summarizing the aforementioned insights, and converting the distortion constraint into a Lagrangian multiplier, we stand to solve the following optimization problem
\begin{align}\label{eq:prac_objective}
    \ell(m) \geq \underset{q(z|x), p(z), \texttt{dec}}{\min} \text{  } \E_{x\sim p_d(x)} \left[ H(q(z|x), p(z))  + \lambda d(x, x^\prime)\right].
\end{align}

It has been noted, that in cases when the distortion metric can be interpreted as an observation model $p(x|z)$ this objective relates to the evidence lower bound (ELBO) of a latent variable model \citep{frey1998bayesian,gregor2016towards,johnston2019computationally,balle2016end,balle2018variational,minnen2018joint}. 
Thus, a lossy compression codec for sequential data can be derived through the respective underlying graphical model. 
A common approach is to construct latent variables by representing each frame $x_i$ as a corresponding latent $z_i$, and using a conditionally independent observation model $p(x|z) = \prod_{i=1}^T p(x_i | z_i)$. However, as can be seen from eq. \ref{eq:prac_objective} even when $z_i$ carries no information, \ie $I(x_i;z_i)=0$, for the entropy codec we are bound to pay a cost for storing the latent frame of $H(q(z|x), p(z))$ bits.
As a result, the bit-rate increases linearly with the number of of observations $T$.
Following we will thus discuss a latent variable model that induces the correct number and temporal location of latent frames as to reduce the overall bit-rate.



\subsection{Latent Stochastic Differential Equations (SDEs)}

Sequential data are often continuous signals that are temporally discretized in order to be efficiently stored on a computer, so a natural representation of such data is a continuous-time process. A stochastic differential equation (SDE) \citep{arnold1974stochastic,oksendal2003stochastic,protter2005stochastic} is a model that describes simultaneously how a continuous signal evolves over time and how much information disperses over time. SDEs can also be interpreted as continuous-time limits of latent variable models~\citep{tzen2019neural}. Given two functions, a drift $f$ and a diffusion $\nu$, an SDE describes the instantaneous change in the state $z(t) : \R \rightarrow \R^{D_z}$ as
\begin{equation}
    d z(t) = f(z(t), t) dt + \nu(z(t), t) dW(t)
\end{equation}
where $W(\cdot)$ is a standard Brownian motion. Sampling from an SDE can be done by providing an initial state $z_0$ at time $t_0=0$, which results in a trajectory $z(\cdot)$ conditioned on the initial value $z(0) = z_0$.
\begin{equation}
    z(\cdot) = \textsc{SDESolve}(z_0, f, \nu, W(\cdot))
\end{equation}
This is a continuous function of $t$ but has to be discretized in order to be represented on a computer, usually on a uniform or adaptive grid, depending on the type of numerical solver scheme being used. While it is possible to directly differentiate through the solver via the reparameterization gradient, memory-efficient gradients can also be computed through an adjoint approach~\citep{li2020scalable}. For the purposes of this work, we can work with any method for solving $z(\cdot)$ and its gradients, and will simply assume we have access to a sufficiently accurate representation of the sample trajectory $z(\cdot)$.

In general, modeling an SDE directly in data space can be difficult as the data is in high dimensions with nontrivial topologies, with marginal distributions $p(z(t))$ that are intractable except for a small class of SDEs.
Instead, we can fit a Latent SDE model~\citep{li2020scalable}, which treats the stochastic process $z$ as a latent variable. Let $\{t_i, x_i\}_{i=1}^T$ be a data sequence with value $x_i$ at time $t_i$. Without loss of generality, we can assume the data lies on a fixed-length interval $[t_0, t_\text{end}]$ with $t_0 = 0$.
If we make the assumption that $x_i$ is conditionally independent of other time values given $z(t_i)$, then this results in a continuous-time hidden Markov model
\begin{equation}
    p(x) = \E_{p(z_0)p(z(\cdot) | z_0)} \left[ \prod_{i=1}^T p(x_i | z(t_i)) \right].
\end{equation}
Note that this model does not require the entire function $z(\cdot)$ but precisely the samples at times $t_1,\dots,t_T$. However, depending on the complexity of $z(\cdot)$, the entire trajectory may be well-approximated by much fewer samples. This motivates our work in learning a data-dependent discretization of $z(\cdot)$ that retains sufficient information regarding the data sequence $x$.

\section{Compression with Latent SDEs}

We propose learning a continuous-time latent representation of sequential data while simultaneously learning how to store a discretization of this representation. For this, a Latent SDE is used as the continuous-time representation while a Latent Temporal Point Process (TPP) determines how to store this representation efficiently. Specifically, for a data sequence $\{t_i, x_i\}_{i=1}^T$, we use the following Latent SDE model with an Ornstein–Uhlenbeck prior
\begin{align}
    z_0 &\sim \mathcal{N}(0, I) \\
    z(\cdot) &= \textsc{SDESolve}(z_0, f_p, \nu, W(\cdot)) \label{eq:model} \\
    x_i &\sim p_\theta(x_i \mid z(t_i)) \;\;\;\text{ for } i=1,\dots,T \label{eq:obs_model}
\end{align}
We set the prior drift as $f_p(z(t), t) \triangleq -\tfrac{1}{2}\nu^2 z_t$, where $\nu \in \R^{D_z}$ denotes the diagonal elements of a constant diffusion and is a learnable parameter; we use $\theta$ to denote free parameters.

This choice of prior SDE results in a stationary process with marginal distributions equal to the standard Normal distribution, acting as a natural extension of the time-independent model to the case where the random variables have dependencies across time. The learning of $\nu$ allows the model to decide on how much mutual information is retained as states diffuse across time.

The approximate inference model $q(z(\cdot) \mid x)$ is an SDE whose drift function depends on the entire sequence $x$. We use a recurrent neural network as a straightforward choice but this can be substituted with any sequence-to-sequence neural network.
\begin{align}
    h(\cdot) &= \textsc{SplineInterp}(\textsc{GRU}(e_\theta(x_i))) \label{eq:hidden_state} \\
    z_0 &\sim \mathcal{N}(z_0 \mid \mu_\theta(h(0)), \sigma^2_\theta(h(0))) \\
    f_q(z(t), t) &\triangleq f_\theta(z(t), h(t)) \\
    z(\cdot) &= \textsc{SDESolve}(z_0, f_q, \nu, W(\cdot))
\end{align}
where $\theta$ denotes the parameters of neural networks with appropriate input and output shapes, and the GRU~\citep{cho2014properties} is bidirectional. This construction allows the drift function to depend on the entire data sequence at all time values. We use a cubic spline interpolation of the output values of the GRU to construct a dense output $h(\cdot)$, so that we can evaluate $f_q$ at any time value.

\paragraph{Lossy Compression} We now have everything necessary for creating a codec for lossy compression. We will store the high-dimensional latent representations with an entropy coder thus
to approximate the average rate for lossy compression, we can estimate the continuous cross entropy 
\begin{equation}\label{eq:lossy_rate}
\begin{split}
    &\varphi_\text{lossy}(x, t_1,\dots,t_T) \triangleq \\
    & \E_{q(z_0 | x) q(z(\cdot) | z_0, x)} \left[ -\log p(z(0), z(t_1), \dots, z(t_T)) \right].
\end{split}
\end{equation}
Since the prior SDE is an Ornstein–Uhlenbeck process, we can compute the probability density function exactly. For a set of points $z_i$ at time values $t_i$ for $i=0,\dots, T$, let $\Delta t_i = t_i - t_{i-1}$, then we have
\begin{equation}
\begin{split}
    &p(z_0, \dots, z_T) =  \\
    & \mathcal{N}(z_0 \mid 0, I) \prod_{i=1}^T \mathcal{N}(z_i \mid z_{i-1} e^{-\tfrac{1}{2}\nu^2 \Delta t_i}, 1 - e^{-\nu^2 \Delta t_i}).
\end{split}
\end{equation}
We train using an estimate of the continuous cross entropy and compute the actual rate after quantizing the latent space post-training.

\paragraph{Lossless Compression} For lossless compression, we will make use of bits-back coding (see \Cref{app:lbitsback} for details). When using bits-back coding the corresponding bit-rate is the KL divergence,
\begin{equation}\label{eq:lossless_rate}
\begin{split}
    &\varphi_\text{lossless}(x, t_1,\dots,t_T) \triangleq \\
    &\E_{q(z_0 | x) q(z(\cdot) | z_0, x)}
    \big[ \log q(z(0),\dots, z(t_T) \mid x) \\
    &\quad\quad\quad\quad\quad\quad\quad\quad - \log p(z(0),\dots, z(t_T)) \big]
\end{split}
\end{equation}
However, the probability density $q$ is generally intractable for general SDEs. Instead, we make use of a pseudo-likelihood~\citep{brouste2014yuima,guidoum2020performing}, which is the exact likelihood under a numerical discretization scheme. Under the Euler discretization, for a set of points $\{t_i, z_i\}$, we have
\begin{equation}
\begin{split}
    &q(z_0,\dots, z_T \mid x) \approx \\ 
    &q(z_0 \mid x) \prod_{i=1}^T \mathcal{N}(z_i \mid z_{i-1} + \Delta t_i f_q(z_{i-1}, t_{i-1}), \Delta t_i \nu^2).
\end{split}
\end{equation}
This is a good approximation to the true density when the drift is nearly constant between each time point. More accurate pseudo-likelihood estimates that use higher order approximations can also be used but would require higher costs to compute. Note that while this pseudo-likelihood is based off an Euler discretization scheme, we need not have simulated $z(\cdot)$ using the same discretization scheme or the same step sizes.

\begin{figure*}
    \centering
    \includegraphics[width=0.85\linewidth]{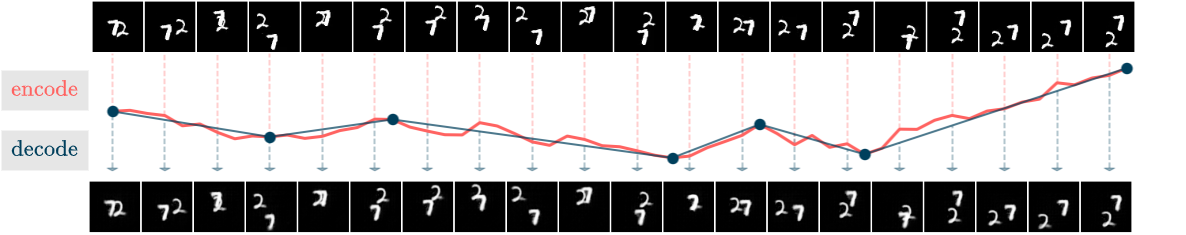} \\
    \caption{
    An illustration of the original uncompressed data sequence (\textit{top}) is encoded into an SDE (\textit{red}) and consequently approximated (\textit{blue}) to decode the datastream into a sequence (\textit{bottom}). Note that we can decode the data stream at any time.}
    \label{fig:pruning_visual}
\end{figure*}

\subsection{Amortized Variational Discretization}\label{sec:amortized_disc}

The main hypothesis of our paper is that we can create good neural compression models even by storing less than $T$ number of samples from $z(\cdot)$ because most sequence data sets have a considerable amount of temporal dependencies. 
Concretely, we can store the knots of a spline that represents an approximation to $z(\cdot)$ which can still retain a high conditional likelihood $p(x_i | z(t_i))$. 
Note that this is a different goal from standard adaptive step size schemes for simulating SDEs; we don't need an accurate representation for $z(\cdot)$ as long as the approximation still retains the necessary information for modeling the data sequence.
As such, we make use of a data-dependent latent temporal point process (TPP), which samples a set of time points $\{\hat{t}_j\}_{j=1}^M$ that indicate the location of the knots. A general overview of TPPs can be found in \citet{daley2003introduction,rasmussen2011temporal}.

We first define a prior TPP using a standard homogeneous Poisson process. Let $\mathcal{T} = (\hat{t}_1, \dots, \hat{t}_M)$ be a set of points on the interval $[0, t_\text{end}]$, then
\begin{equation}\label{eq:prior_tpp}
    p(\mathcal{T}) = \lambda^M e^{-t_\text{end}\lambda}
\end{equation}
where the intensity parameter $\lambda$ is equal to the average number of points per unit time. We treat this as a hyperparameter and keep it fixed throughout training. Note that this is a very simple prior with a probability that only depends on the number of points, allowing our inference model the complete freedom to dictate the location of the points.

While TPPs are often characterized by an intensity function~\citep{rasmussen2011temporal}, we take the simpler approach and directly parameterize distributions over the inter-event times, $\Delta \hat{t}_j$ such that $\hat{t}_j = \sum_{k=1}^j \Delta \hat{t}_k$, similar to \citet{shchur2020intensity}. This parameterization provides more fine-grained control over the spacing between samples.
The probability of sampling a set of $M$ points from the interval $[0, t_\text{end}]$ is
\begin{equation}\label{eq:posterior_tpp}
\begin{split}
    q(\mathcal{T} \mid x) = &q(\Delta \hat{t}_1 \mid x) q(\Delta \hat{t}_2 \mid \hat{t}_1, x) \cdots \\
    &\quad q(\Delta \hat{t}_M \mid \hat{t}_{M-1}, x) \mathbb{P}(\Delta \hat{t}_{M+1} > t_\text{end} - \hat{t}_M \mid x).
\end{split}
\end{equation}
The last term takes into account the probability of not seeing another point between the last sampled point $\hat{t}_M$ and the end of the interval, which can be obtained from the cumulative distribution function (CDF). We condition the distributions $q(\Delta \hat{t}_j \mid \hat{t}_{j-1}, x)$ on the data sequence by depending on the hidden state at time $\hat{t}_{j-1}$ from \cref{eq:hidden_state}.
\begin{equation}
\begin{split}
    &q(\Delta \hat{t}_j \mid \hat{t}_{j-1}, x) \triangleq \\
    &\text{SoftplusLogistic}(\Delta \hat{t}_j \mid \mu_\phi(h(\hat{t}_{j-1}), \hat{t}_{j-1}), \\
    &\quad\quad\quad\quad\quad\quad\quad\quad\quad\;\; s_\phi(h(\hat{t}_{j-1}), \hat{t}_{j-1}))
\end{split}
\end{equation}
where $\phi$ denotes the free parameters of this discretization model. 
Essentially, a neural network outputs the mean and variance of a logistic distribution which is then transformed by through a  softplus function. 
This has the benefit of having an exact CDF while the density function is available through a change of variables. Furthermore, we truncate this distribution to control the maximum change in time, as this helps stabilize training. The exact equations can be found in \Cref{app:q_tpp}.

We then represent this discretized trajectory by taking the conditional expectation under the prior SDE, which for the Ornstein-Uhlenbeck process, is computationally equivalent to a linear spline interpolation. Conditioned on the two end points $z(t_0)$ and $z(t_\text{end})$, we have
\begin{align}
\begin{split}
    &\hat{z}(\cdot) \triangleq \E_{p(z(\cdot) | z(t_0), z(\hat{t}_1), \dots, z(\hat{t}_M), z(t_\text{end}))}[ z(\cdot) ] = \\
    &\textsc{SplineInterp}(z(t_0), z(\hat{t}_1), \dots, z(\hat{t}_M), z(t_\text{end})), \\
    &\quad\quad\quad\quad\quad\quad\quad\quad \text{ where } \hat{t}_1, \dots, \hat{t}_M \sim q(\mathcal{T} \mid x),
\end{split}
\end{align}
which can then be decoded at arbitrary time values to reconstruct the data sequence. We refer to this as a \methodname for \emph{variationally discretized} SDE.

With this, we can now construct the full training objective. For a data sequence $\{t_i, x_i\}_{i=1}^T$,
\begin{equation}\label{eq:training_obj}
\begin{split}
    \mathcal{L}(\{t_i, x_i\}_{i=1}^T) &= \underbrace{\E_{\hat{z}(\cdot)} \left[ - \sum_{i=1}^T \log p(x_i \mid \hat{z}(t_i)) \right]}_{\text{reconstruction}} \\
    &+ \underbrace{\varphi(\hat{t}_1,\dots, \hat{t}_M)}_{\text{bits for encoding $\{z(\hat{t}_i)\}$}} \\
    &+ \underbrace{\mathcal{D}_\text{KL}(q(\hat{t}_1,\dots,\hat{t}_M \mid x) \;\|\; p(\hat{t}_1,\dots,\hat{t}_M))}_{\text{bits for encoding $\{\hat{t}_i\}$}}
\end{split}
\end{equation}
where $\varphi$ is either \cref{eq:lossy_rate} or \cref{eq:lossless_rate} depending on the compression method. 

With a learned discretization model, we must also take into account storage of the discretization grid itself, $\T$. Since it is a one-dimensional variable we can use a REC compression scheme,
which has a theoretical rate equal to the divergence $\mathcal{D}_{\text{KL}}$. This does slightly increase the total compression rate, but this is very minuscule when compared to the bit rate from compressing the latent variables $z(t_j)$. 

\paragraph{Stochastic gradients}

Since every sample from a temporal point process is a set with random cardinality, the discrete nature of this prohibits the use of unbiased reparameterization gradients. Concretely, a slight change in parameter space can result in a different number of samples, so the probability density functions in \cref{eq:prior_tpp,eq:posterior_tpp} are not continuous with respect to parameters of the discretization model $\phi$. Therefore, we disable gradients through time samples $\{\hat{t}_j\}_{j=1}^M$ and we train the discretization model using \textsc{Reinforce} gradients~\citep{williams1992simple} while the rest are trained with reparameterization gradients through $z(\cdot)$ and \textsc{SDESolve}.
\begin{equation}
\begin{split}
    \nabla_{\theta, \phi} \mathcal{L} =& \E_{z(\cdot), \{t_j\}_{j=1}^M} \big[ \\
    &\mathcal{L}(\{t_i, x_i\}_{i=1}^T) \nabla_\phi \log q(\{\hat{t}_j\}_{j=1}^M \mid x) \\
    &+ \nabla_\theta \mathcal{L}(\{t_i, x_i\}_{i=1}^T) \big]
\end{split}
\end{equation}
Note that the \textsc{Reinforce} gradient does not backpropagate through the hidden state or any parameters outside of the discretization model. 
In practice, \textsc{Reinforce} gradients can have high variance, and we found that introducing the batch averaged loss value as a simple baseline works sufficiently well. 

\paragraph{Two-stage training} At initialization, the model has poor reconstruction and the latent process does not yet contain useful information. As such, the discretization model realizes that it can use a very sparse grid without affecting the distortion while significantly reducing rate. This often results in the model becoming stuck at this poor local optimum without the ability to recover. To remedy this, we use a two-stage training process, where in the first stage we only train the SDE portion of the model and use the full discretization of the data sequence, setting $\mathcal{T} = \{t_1,\dots, t_T\}$. In the second stage, we then switch to sampling from the discretization model $q(\T \mid x)$ and train both the discretization and the SDE model jointly using \cref{eq:training_obj}.

\subsection{Local and Global Latent Dimensions}\label{sec:pruning}

With our particular choice of prior, the model can learn different dynamics for each latent dimension due to different values of $\nu$ in \cref{eq:model}. We find that in trained models, multiple dimensions have a value of $\nu$ that is very close to zero. This indicates that both the prior drift and diffusion are nearly zero, leading to a constant trajectory. We thus find these trajectories are well approximated by the initial value itself. As such, when the trained value $\nu \leq 0.001$, then during evaluation, we set the entire trajectory to be constant $z(\cdot) = z_0$. As a result, we can significantly save on rate in the lossy compression setting by storing only the initial value for these dimensions, effectively pruning a majority of the latent states, without sacrificing compression quality (\Cref{fig:pruning_visual}). 

\section{Related Work}

\begin{figure*}
    \centering
    \begin{subfigure}[b]{0.28\linewidth}
         \centering
         \includegraphics[width=\linewidth]{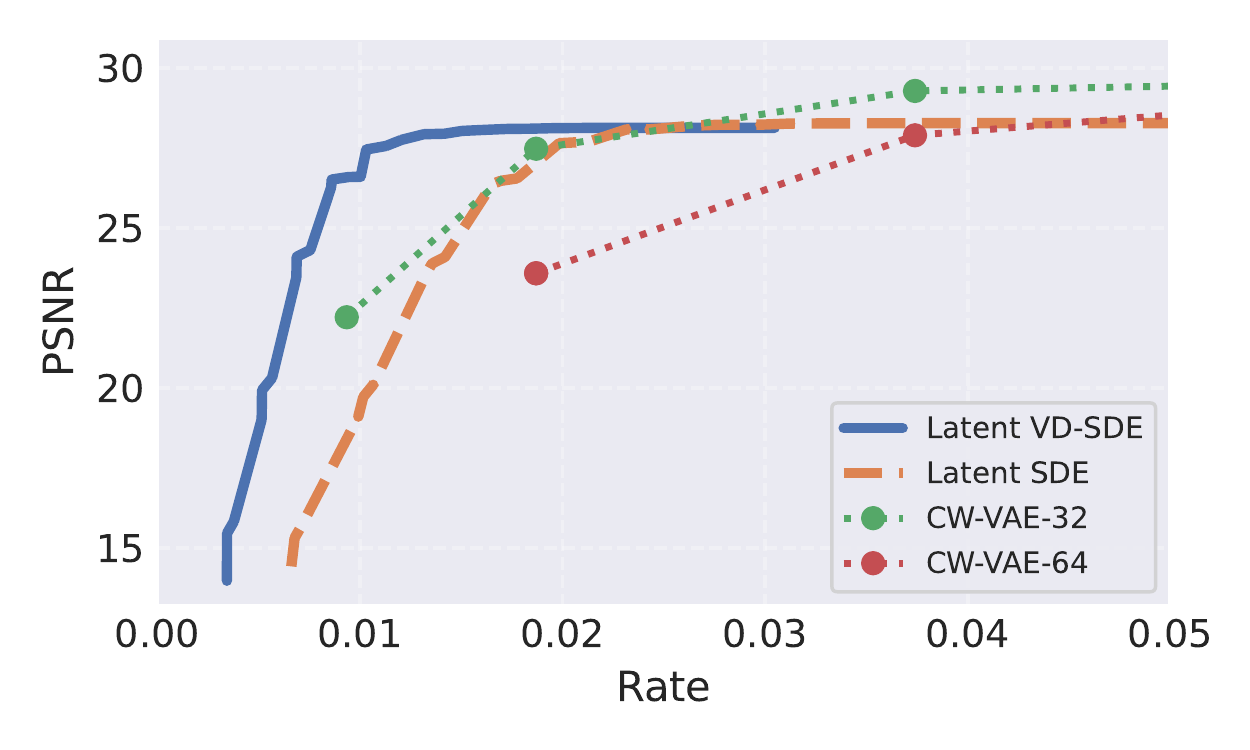}
         \caption*{GQN Mazes}
    \end{subfigure}
    \begin{subfigure}[b]{0.28\linewidth}
         \centering
         \includegraphics[width=\linewidth]{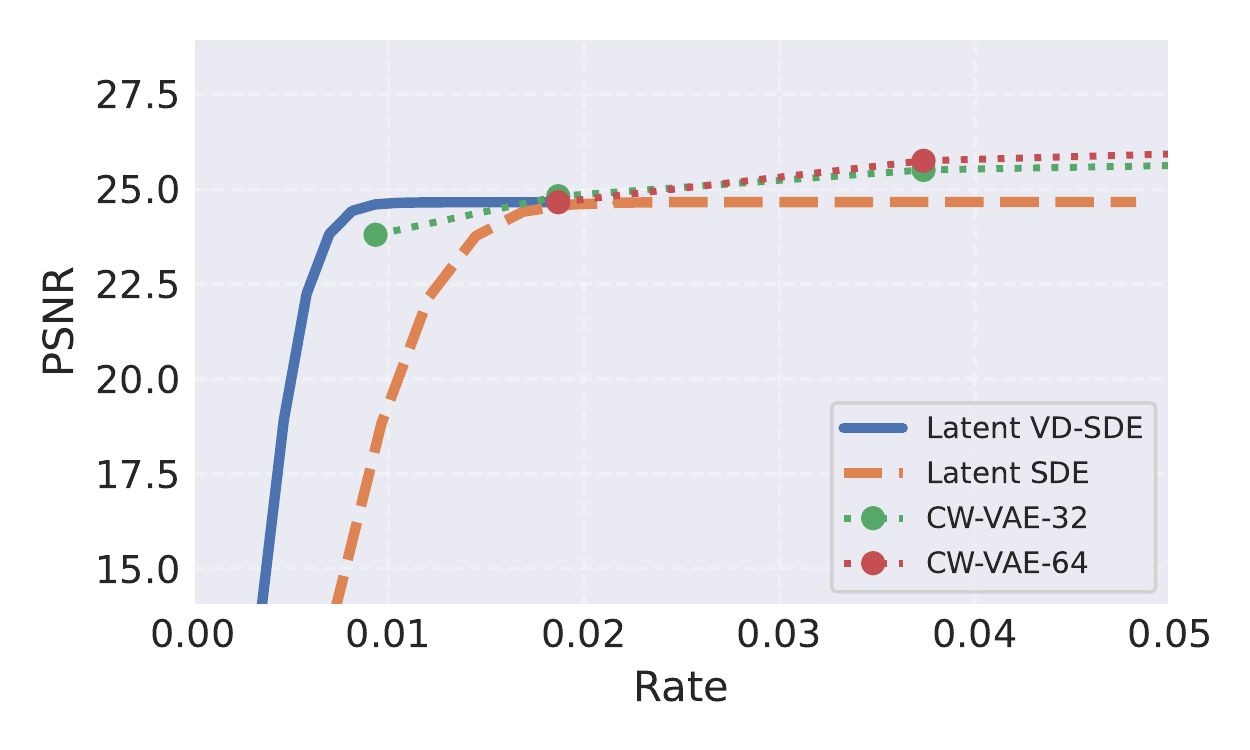}
         \caption*{MineRL}
    \end{subfigure}
    \begin{subfigure}[b]{0.28\linewidth}
         \centering
         \includegraphics[width=\linewidth]{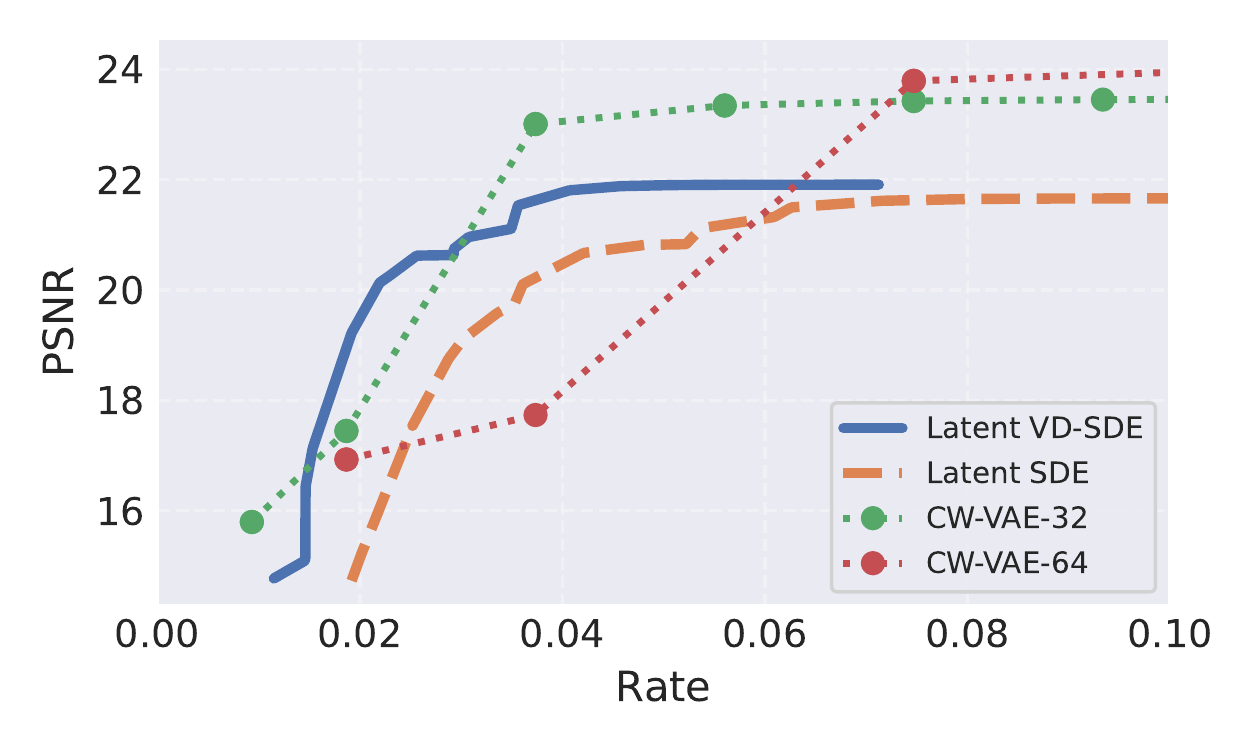}
         \caption*{MovingMNIST}
    \end{subfigure}
    \caption{Rate-distortion curves for video sequence data sets. A Latent SDE model performs on par with CW-VAE models. However, when optimizing time discretization our model outperforms CW-VAE in the low rate regime. }
    \label{fig:video_results}
\end{figure*}

\vspace{-0.7em}
\paragraph{Continuous-time modeling}
Modeling differential equations with deep neural networks~\citep{chen2018neural} has gained traction in the machine learning community for being memory- and data-efficient. Using these models in the latent space~\citep{rubanova2019latent,de2019gru,li2020scalable,rempe2020caspr} provides methods to extract continuous-time representations from sequence data. These approaches have mainly been used for accurately modeling irregularly-sampled time series~\citep{rubanova2019latent,de2019gru,morrill2021neural}. Here we motivate a new application in efficiently compressing sequences that are densely sampled.

While some discussion exists around the role of discretization in simulating ordinary differential equations~\citep{ott2020resnet,queiruga2020continuous,kidger2021hey}, these are mainly targeted at retaining numerical accuracy of the simulations. 
In contrast, we assume the simulations are sufficiently accurate and discuss the use of discretization only for the task of compression, \ie storing on a computer. 
Furthermore, existing adaptive step size algorithms for stochastic differential equations~\citep{burrage2004adaptive,lamba2007adaptive} are based on local heuristics, while our variational discretization model is conditioned on an entire data sequence and designed to work for all samples from a stochastic process. 
These combined allow the discretization model to create a much coarser discretization grid while maintaining accurate data reconstruction. While the use of splines has been combined with neural differential equation models~\citep{daulbaev2020interpolation,kidger2020neural} in order to save memory or compute costs, we make use of these in a stochastic context and differentiate through the spline interpolation for learning.

\vspace{-0.7em}
\paragraph{Latent temporal point processes}
While temporal point processes based on neural networks have been explored quite extensively~\citep{jia2019neural,chen2020neural}, the majority of recent works are focused on fitting point processes to data~\citep{shchur2021neural}. We tackle the problem of using TPP models in a latent setting, where sampling-based optimization is required for learning. 
This can pose difficulties as the density function of temporal point processes is not reparameterizable without fixing the number of points~\citep{shchur2020fast,chen2020learning}. In particular, \citet{li2018learning} explores the use of latent TPPs for reinforcement learning and uses policy gradients for optimization, while \citet{chen2020learning} notes that if the objective is continuous, then the reparameterization gradient can be used as well. As the density function shows up in our objective, we defer to using policy gradients with a simple control variate for increasing training stability.

\vspace{-0.7em}
\paragraph{Compression of sequential data}
Compression methods for sequential data depend on the specific data modality, e.g. video, weather observations, motion capture, radar etc.
Typically, a modality specific function encodes observations or temporal differences in observations into a latent space, and only this latent representation is then compressed losslessly with a model-based entropy coder. However, existing models and compression algorithms are mostly dependent on the data modality.

For example classic video compression methods, such as HEVC/H.265, AVC/H.264 and VP9, compress the residual of a frame prediction based two or more neighboring frames with different models, with frames categorized into I-, P-, and B-frames. Furthermore, the locations of these frames are not learned but predefined based on a user-provided framerate.
As a consequence of this approach it is not trivial to convert videos from one frame rate to another. 
More recently, machine learning has been introduced into the video compression pipeline in order improve the complexity of modeling frame residuals \citep{rippel2019learned,agustsson2020scale,chen2017deepcoder,wu2018video,djelouah2019neural,golinski2020feedback,lu2020end,yang2020hierarchical} and the entropy coder \citep{gregor2016towards,habibian2019video,lu2019dvc,yang2020learning}. 
Even though, the idea of assigning variable bitrates across space e.g. \citep{rippel2019learned} is common, to the best of our knowledge, there is no method that considers amortized temporal discretization in either the classic or neural compression literature.

Another example are sequences of point clouds. At the highest difficulty, 
LiDAR sequences in self-driving cars produce up to 45 GB per hour per sensor~\citep{choi2016millimeter}.
Though the process being captured is continuous-time signal, often this is discretized according to a regular grid both spatially and temporally. 
The spatial location of each point cloud (i.e. frame) is often discretized and either treated like an image \citep{mekuria2016design,tu2016compressing,tu2019real,tu2019point} or an octree \citep{biswas2020muscle}, which is subsequently compressed with a learned entropy model. Generally, these domain-specific approaches can all be combined with a learnable discretization model.

\begin{figure*}
    \centering
    
    \raisebox{2em}{\rotatebox[origin=t]{90}{\small Original}}%
    \includegraphics[width=0.095\linewidth, trim=150px 0 150px 0, clip]{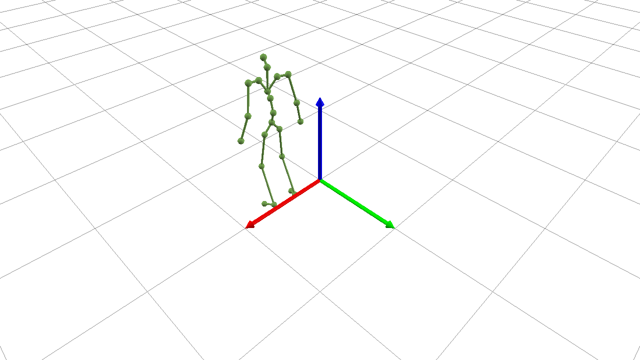}%
    \includegraphics[width=0.095\linewidth, trim=150px 0 150px 0, clip]{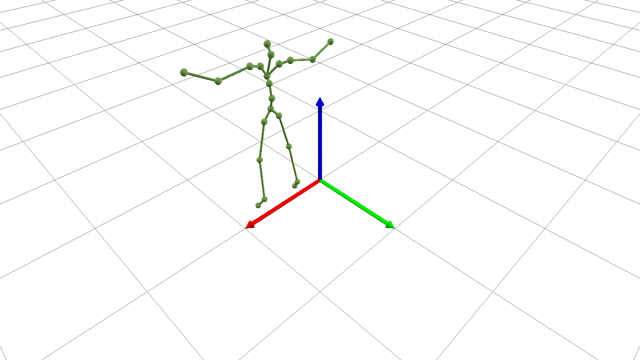}%
    \includegraphics[width=0.095\linewidth, trim=150px 0 150px 0, clip]{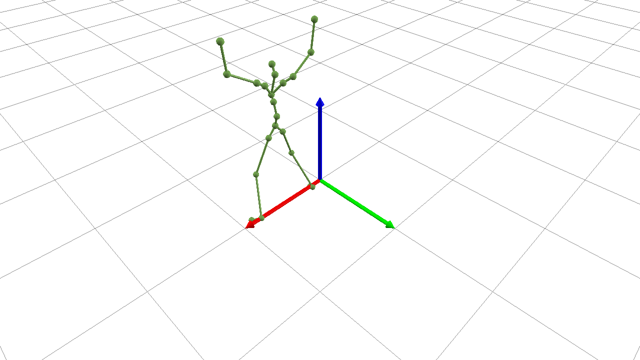}%
    \includegraphics[width=0.095\linewidth, trim=150px 0 150px 0, clip]{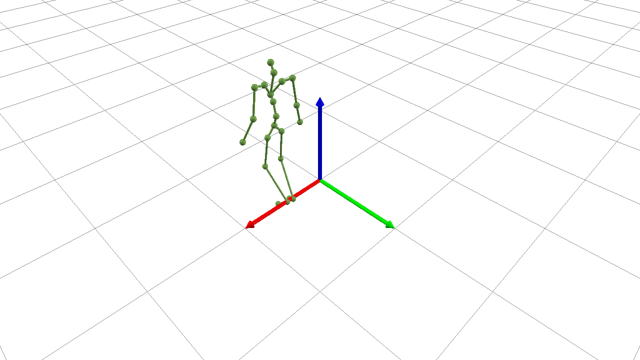}%
    \includegraphics[width=0.095\linewidth, trim=150px 0 150px 0, clip]{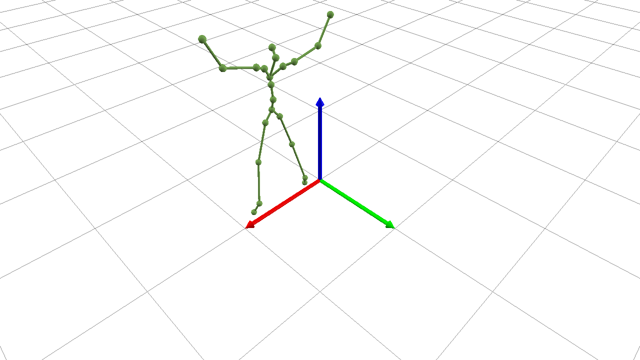}%
    \includegraphics[width=0.095\linewidth, trim=150px 0 150px 0, clip]{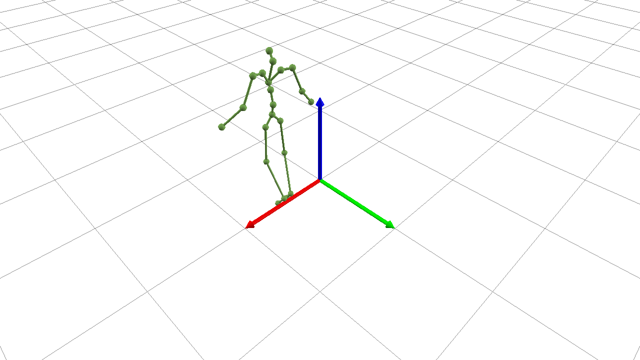}%
    \includegraphics[width=0.095\linewidth, trim=150px 0 150px 0, clip]{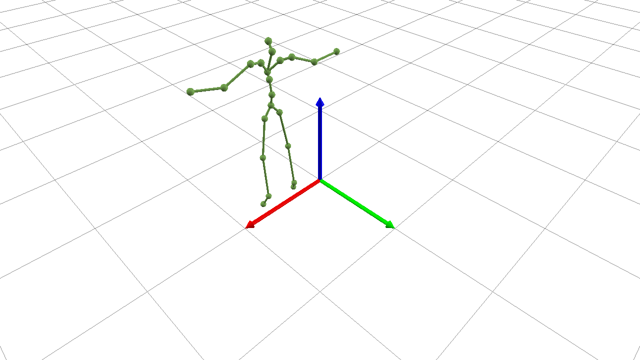}\\
    \raisebox{2em}{\rotatebox[origin=t]{90}{\small Compressed}}%
    \includegraphics[width=0.095\linewidth, trim=150px 0 150px 0, clip]{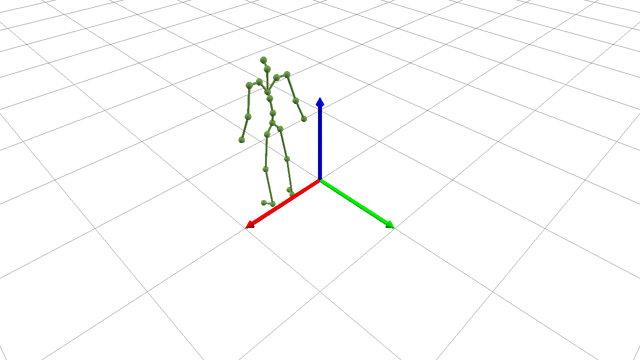}%
    \includegraphics[width=0.095\linewidth, trim=150px 0 150px 0, clip]{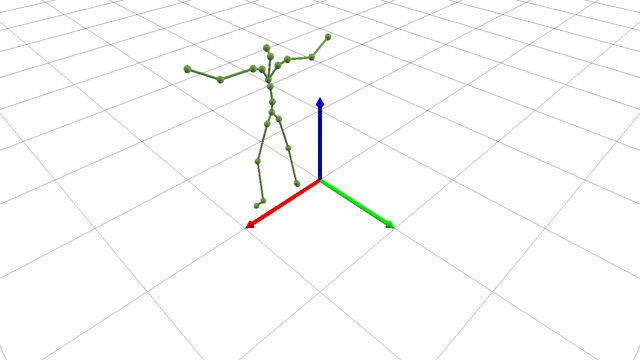}%
    \includegraphics[width=0.095\linewidth, trim=150px 0 150px 0, clip]{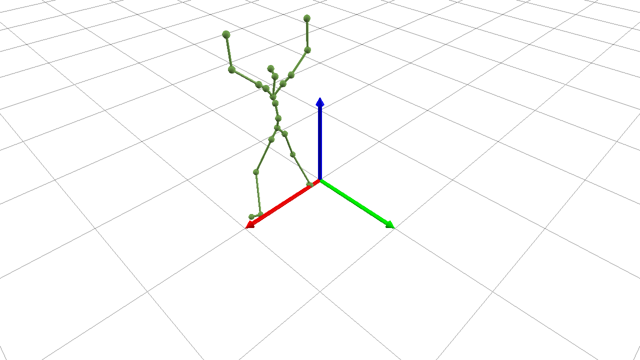}%
    \includegraphics[width=0.095\linewidth, trim=150px 0 150px 0, clip]{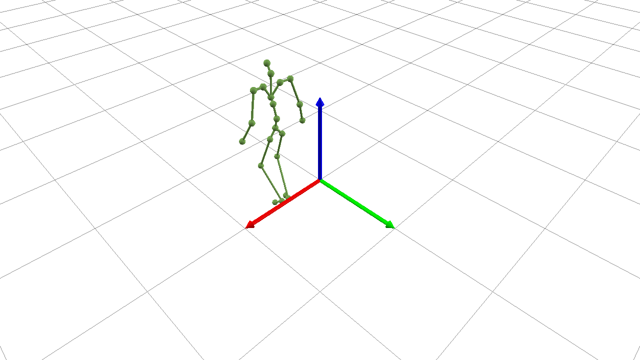}%
    \includegraphics[width=0.095\linewidth, trim=150px 0 150px 0, clip]{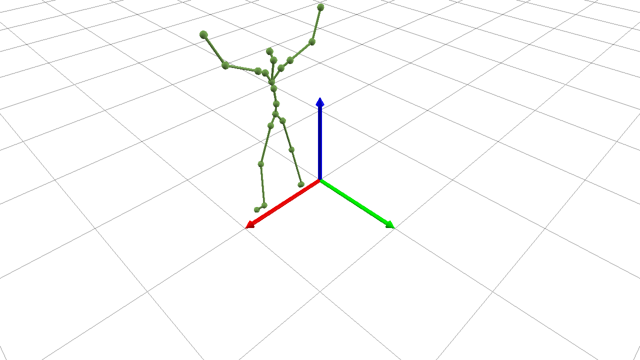}%
    \includegraphics[width=0.095\linewidth, trim=150px 0 150px 0, clip]{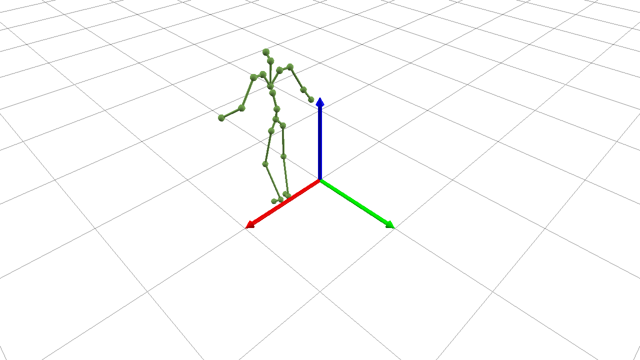}%
    \includegraphics[width=0.095\linewidth, trim=150px 0 150px 0, clip]{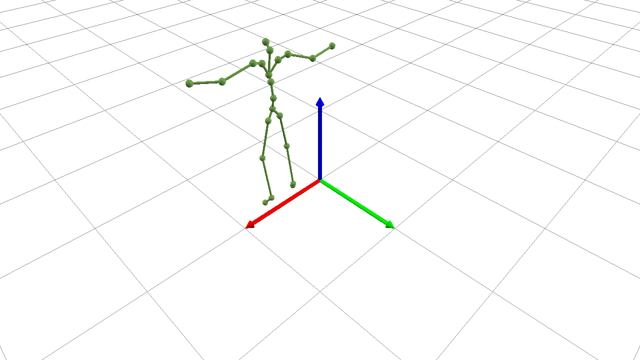}\\
    
    \vspace{0.5em}
    
    \raisebox{2em}{\rotatebox[origin=t]{90}{\small Original}}%
    \includegraphics[width=0.095\linewidth, trim=150px 0 150px 0, clip]{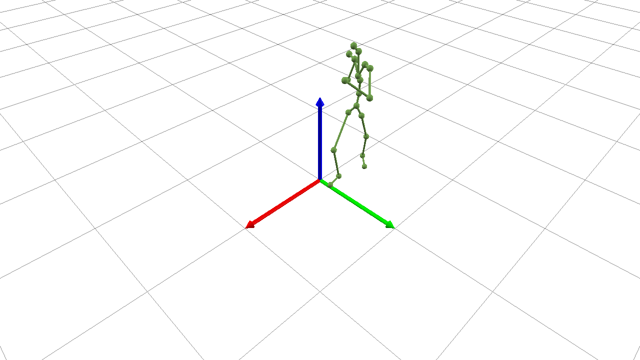}%
    \includegraphics[width=0.095\linewidth, trim=150px 0 150px 0, clip]{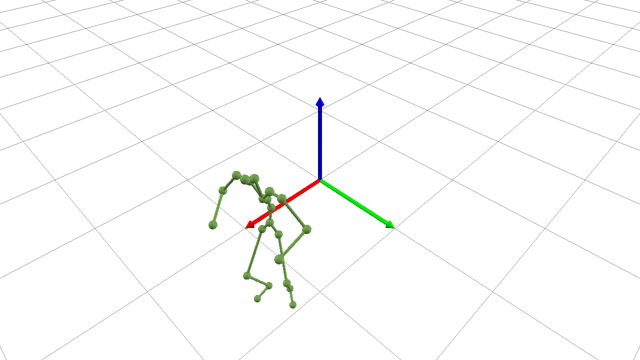}%
    \includegraphics[width=0.095\linewidth, trim=150px 0 150px 0, clip]{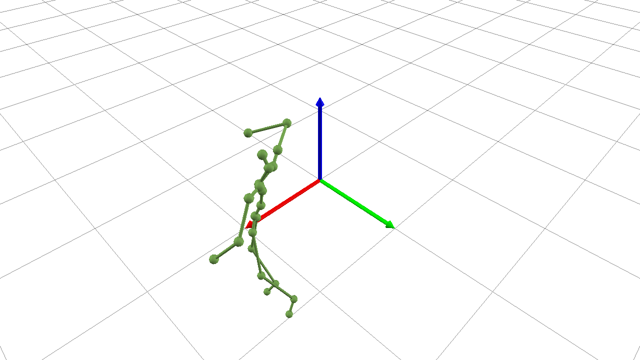}%
    \includegraphics[width=0.095\linewidth, trim=150px 0 150px 0, clip]{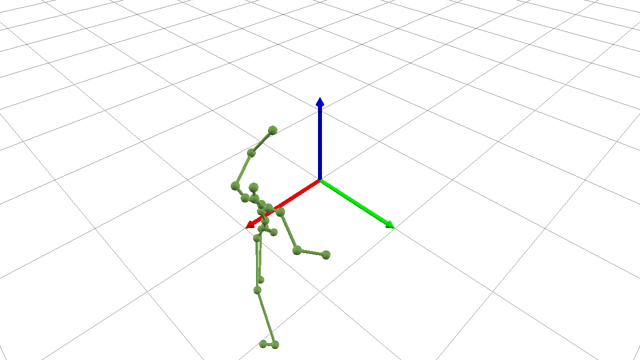}%
    \includegraphics[width=0.095\linewidth, trim=150px 0 150px 0, clip]{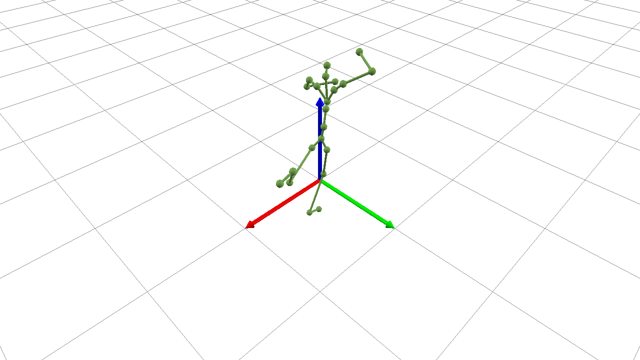}%
    \includegraphics[width=0.095\linewidth, trim=150px 0 150px 0, clip]{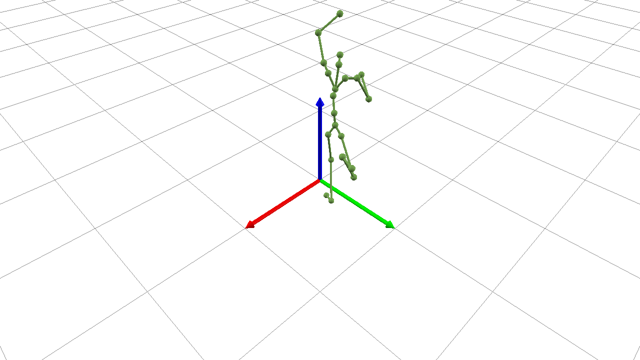}%
    \includegraphics[width=0.095\linewidth, trim=150px 0 150px 0, clip]{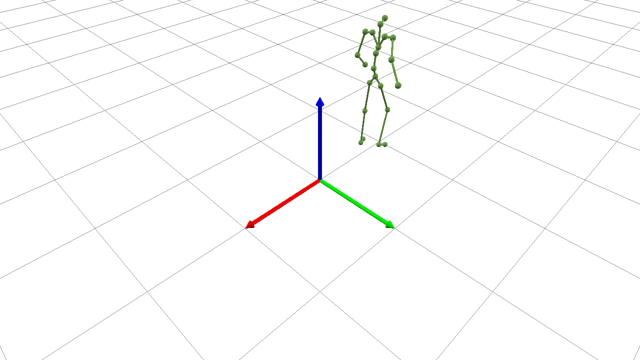}\\
    \raisebox{2em}{\rotatebox[origin=t]{90}{\small Compressed}}%
    \includegraphics[width=0.095\linewidth, trim=150px 0 150px 0, clip]{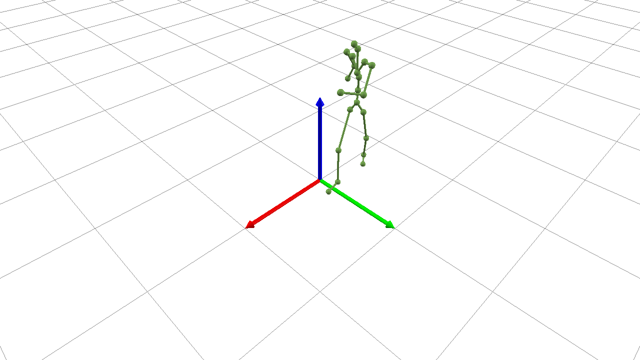}%
    \includegraphics[width=0.095\linewidth, trim=150px 0 150px 0, clip]{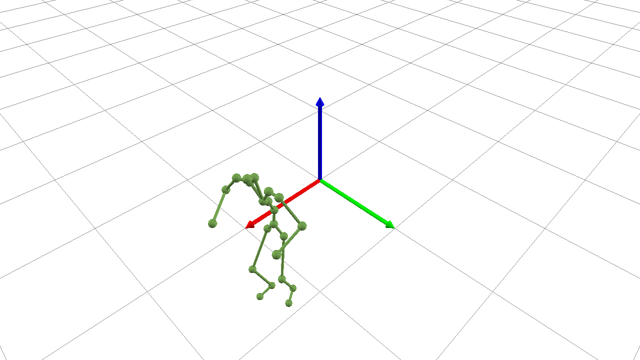}%
    \includegraphics[width=0.095\linewidth, trim=150px 0 150px 0, clip]{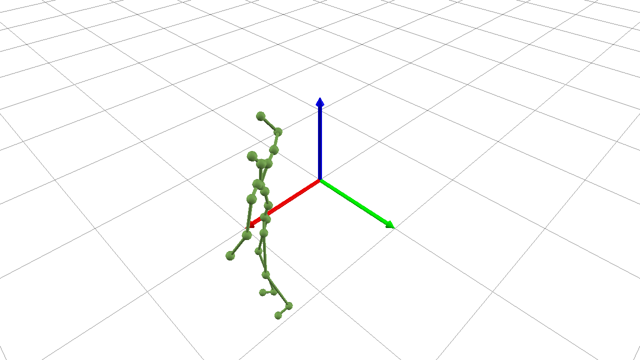}%
    \includegraphics[width=0.095\linewidth, trim=150px 0 150px 0, clip]{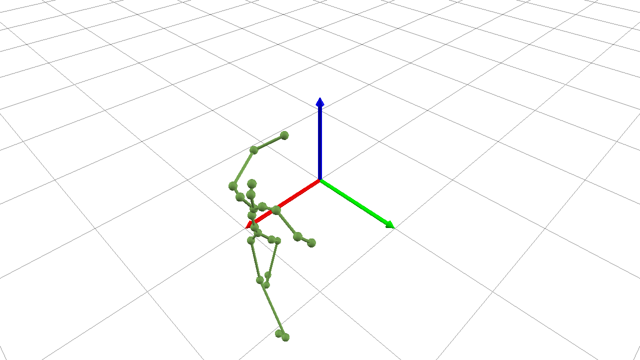}%
    \includegraphics[width=0.095\linewidth, trim=150px 0 150px 0, clip]{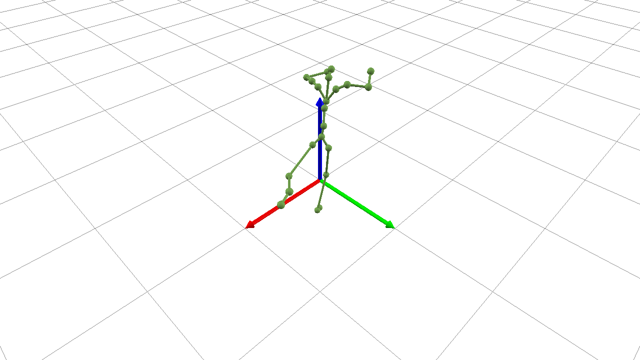}%
    \includegraphics[width=0.095\linewidth, trim=150px 0 150px 0, clip]{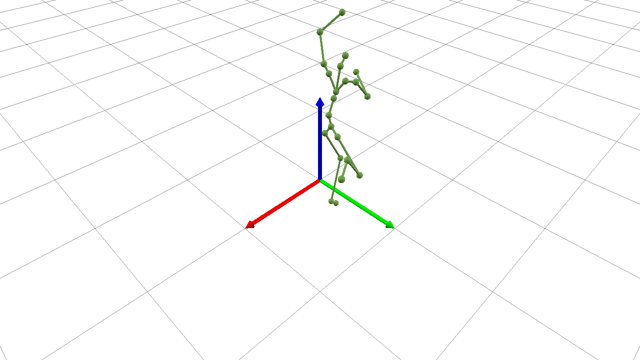}%
    \includegraphics[width=0.095\linewidth, trim=150px 0 150px 0, clip]{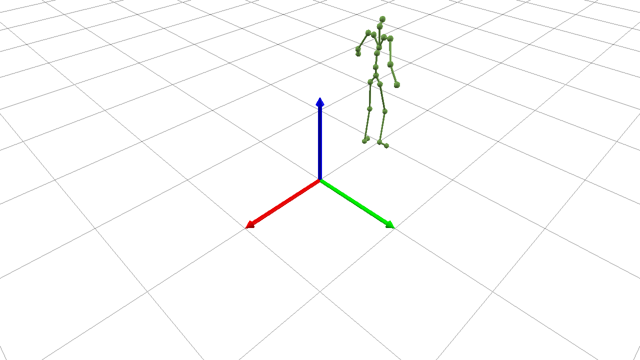}\\

    \caption{Compressed sequences with around 98\% average reduction in communication cost. Main overall movements are all captured by the model for both smooth (\textit{top}) and erratic (\textit{bottom}) sequences.}
    \label{fig:mocap_visual}
\end{figure*}

\section{Experiments}

We experiment on data sets involving video and motion capture sequences. 
In particular, we make use of data sets of video sequences preprocessed by \citet{saxena2021clockwork}. 
These include the the MineRL Navigate data set~\citep{guss2019minerl} (scenes as players traverse through the world of Minecraft), GQN Mazes~\citep{eslami2018neural} (frames of an agent traversing through a randomized maze), and Moving MNIST~\citep{srivastava2015unsupervised} (frames with 2 digits moving with variable velocity).
We also experiment with motion capture data sets, where there exist an underlying continuous motion.
For this, we make use of publicly available data sets from CMU motion capture~\citep{de2009guide} and a large set of aggregated motion capture data, AMASS~\citep{mahmood2019amass}. 


We compare against the Clockwork VAE~(CW-VAE; \citet{saxena2021clockwork}), a state-of-the-art latent sequence model. 
This model uses multiple levels of latent sequences, each having its own fixed discretization interval. Although this efficiently handles different temporal abstractions, every level of latent variables need to be stored independently. 
For more flexibility, we manually train multiple CW-VAE models with varying number of latent frame dimensionality, denoted CW-VAE-XX. 

When compressing latent variables $z$, we quantize them using maximum entropy discretization~\citep{cover2012elements}, which are often used similarly in prior works~(\eg \citet{townsend2019practical}), and sweep over a range of precision values to generate rate-distortion plots. With amortized discretization grids, we must also compress the set $\T$. However, as these are simple 1D random variables, we can make use of relative entropy coding schemes~\citep{li2018strong,flamich2020compressing,flamich2022fast,theis2021algorithms} which adds an additional cost of $\mathcal{D}_\text{KL}(q(\hat{t} | x) || p(\hat{t}))$ to the rate. Similar to contemporary works (e.g. \citet{theis2022lossy}), we report estimated rates for relative entropy coding, but the reported values should be representative of the actual rate, up to a negligible additive constant (see \Cref{app:relative_entropy_coding}). 

\begin{table}
\centering
\caption{Lossless rates for compressing the video data sets, with estimated standard deviation.}
\label{tab:lossless}
\ra{1.2}
\setlength{\tabcolsep}{2.5pt}
\resizebox{\linewidth}{!}{%
\begin{tabular}{@{} l c c c c c c c c @{}}\toprule
   &
  \multicolumn{2}{c}{\bf Moving MNIST} & &
  \multicolumn{2}{c}{\bf GQN Mazes} & &
  \multicolumn{2}{c}{\bf MineRL} \\
  Model & 
  KL & Total Rate & & 
  KL & Total Rate & & 
  KL & Total Rate \\
\midrule
Latent SDE & 
171.75{\tiny$\pm$1.63} & 0.810{\tiny$\pm$0.002} & &
348.19{\tiny$\pm$9.34} & 13.300{\tiny$\pm$0.025} & &
223.41{\tiny$\pm$9.01} & 16.435{\tiny$\pm$0.008} \\
Latent VD-SDE &  
\textbf{126.55{\tiny$\pm$2.44}} & \textbf{0.798{\tiny$\pm$0.003}} & &
\textbf{240.39{\tiny$\pm$3.27}} & \textbf{13.230{\tiny$\pm$0.003}} & &
\textbf{161.50{\tiny$\pm$0.36}} & \textbf{16.363{\tiny$\pm$0.003}} \\
\bottomrule
\end{tabular}
}
\vspace{-1em}
\end{table}

\subsection{Video Compression}

\vspace{-0.3em}
\paragraph{Experiment details}
We use sequence lengths of $100$ frames at $64\times 64$ resolution. We generally stick to the same hyperparameters as \citet{saxena2021clockwork}: all models are trained using Adam with a fixed learning rate of $0.0003$ except MineRL which used a learning rate of $0.0001$; the observation model (\eqref{eq:obs_model}) is isotropic Normally distributed with a fixed standard deviation of $0.1$ for all data sets except MineRL which uses $0.3$. 
All of our models are trained for $150000$ iterations. We sweep over the prior TPP's intensity value $\lambda$ between 50\% to 100\% of the total sequence length.

\vspace{-0.7em}
\paragraph{Rate-distortion curves} In \Cref{fig:video_results} we show the rate-distortion curves for a Latent SDE baseline and our improved version Latent VD-SDE with learned discretization. 
Compared to the baseline Latent SDE model, our learned discretization model successfully reduces bit rate, by around 30\% to 60\% depending on the data set, while retaining very similar PSNR compared to the full Latent SDE model. As our model (around 5 million parameters) is much smaller than the CW-VAE (which has around 28 million parameters), it obtains slightly lower optimal PSNR values at the high rate regime before any discretization of the latent space. Nevertheless, the rates that we obtain were automatically learned by the network, both as a result of the learned discretization and the pruning of latent dimensions, whereas for the baseline CW-VAE, low rates could only be obtained by sweeping over many hyperparameter values such as the size of the latent dimension.

\begin{figure*}
    \centering
    \begin{subfigure}[b]{0.32\linewidth}
         \centering
         \includegraphics[width=\linewidth, trim=0 526px 0 0, clip]{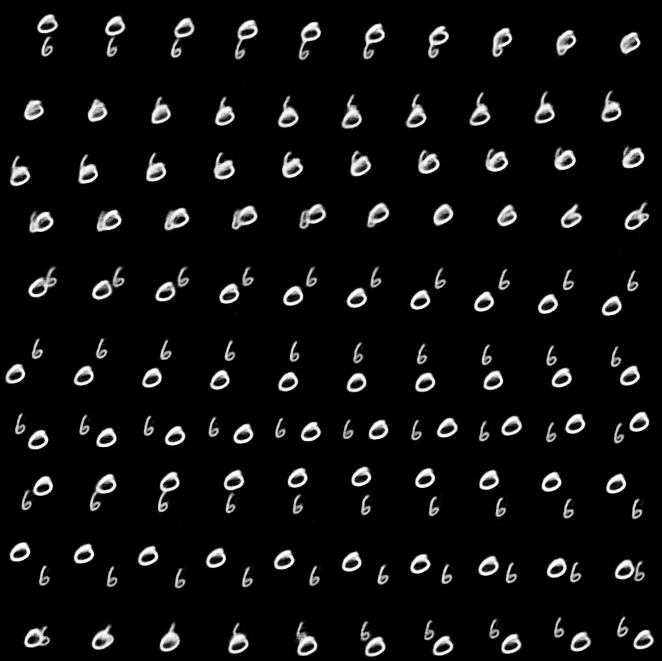}
         \caption{Global Reference}
    \end{subfigure}
    \begin{subfigure}[b]{0.32\linewidth}
         \centering
         \includegraphics[width=\linewidth, trim=0 526px 0 0, clip]{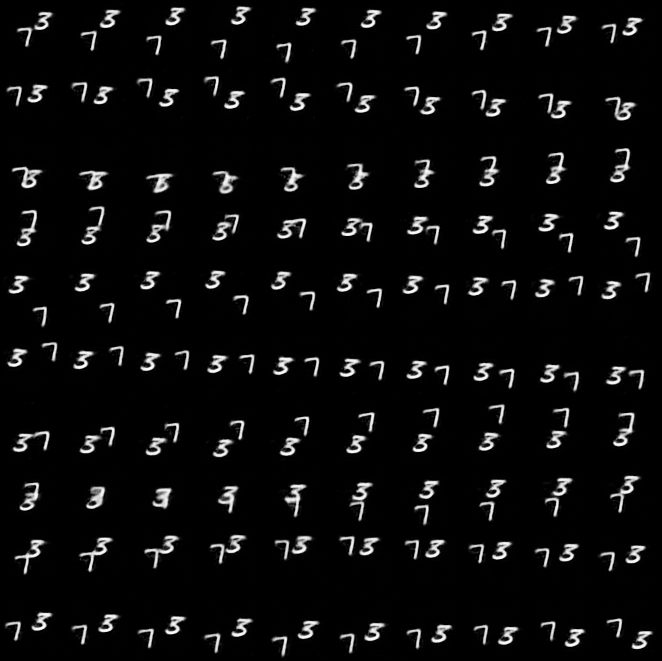}
         \caption{Local Reference}
    \end{subfigure}
    \begin{subfigure}[b]{0.32\linewidth}
         \centering
         \includegraphics[width=\linewidth, trim=0 526px 0 0, clip]{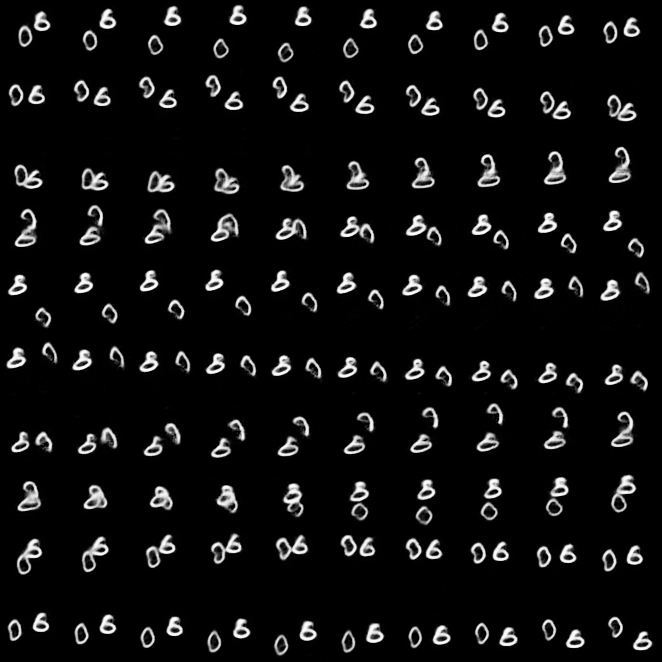}
         \caption{Combined Sequence}
    \end{subfigure}
    \caption{Combining the local latent states with the global latent states of another sequence. The digits are modified to the global reference while the positions remain the same as the local reference.}
    \label{fig:disentanglement}
\end{figure*}

\vspace{-0.7em}
\paragraph{Global vs local latent dimensions} In some cases, we can visually identify what properties the model has allocated to the global latent dimensions. In the case of moving MNIST digits, the global representation contains details regarding the character class while the local time-varying representations denote spatial location. See \Cref{fig:disentanglement}.

\vspace{-0.7em}
\paragraph{Effect of pruning global latent dimensions} The results shown in \Cref{fig:video_results} include the pruning procedure discussed in \Cref{sec:pruning}. This allows us to use a larger latent dimension, as we can significantly reduce most of the cost at compression time. In contrast, a discrete-time model such as the CW-VAE baseline must be manually trained with varying latent dimensions, as it is unknown a priori how many latent dimensions is required to model a particular data set. Since we prune only dimensions that are effectively numerically constant, there is almost no change in the distortion. A qualitative comparison is shown in \Cref{fig:pruning_visual}.

\begin{figure*}
    \centering
    \begin{subfigure}[b]{0.28\linewidth}
         \centering
         \includegraphics[width=\linewidth]{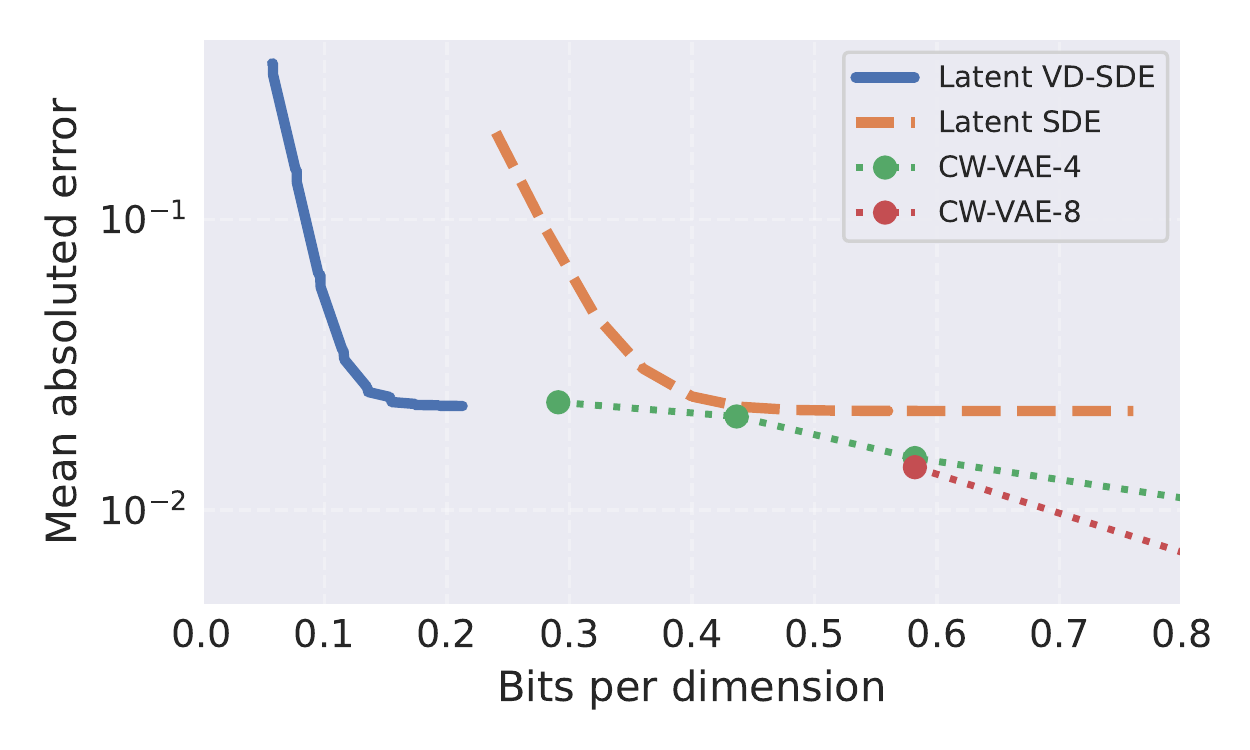} 
         \caption*{CMU}
    \end{subfigure}
    \begin{subfigure}[b]{0.28\linewidth}
         \centering
         \includegraphics[width=\linewidth]{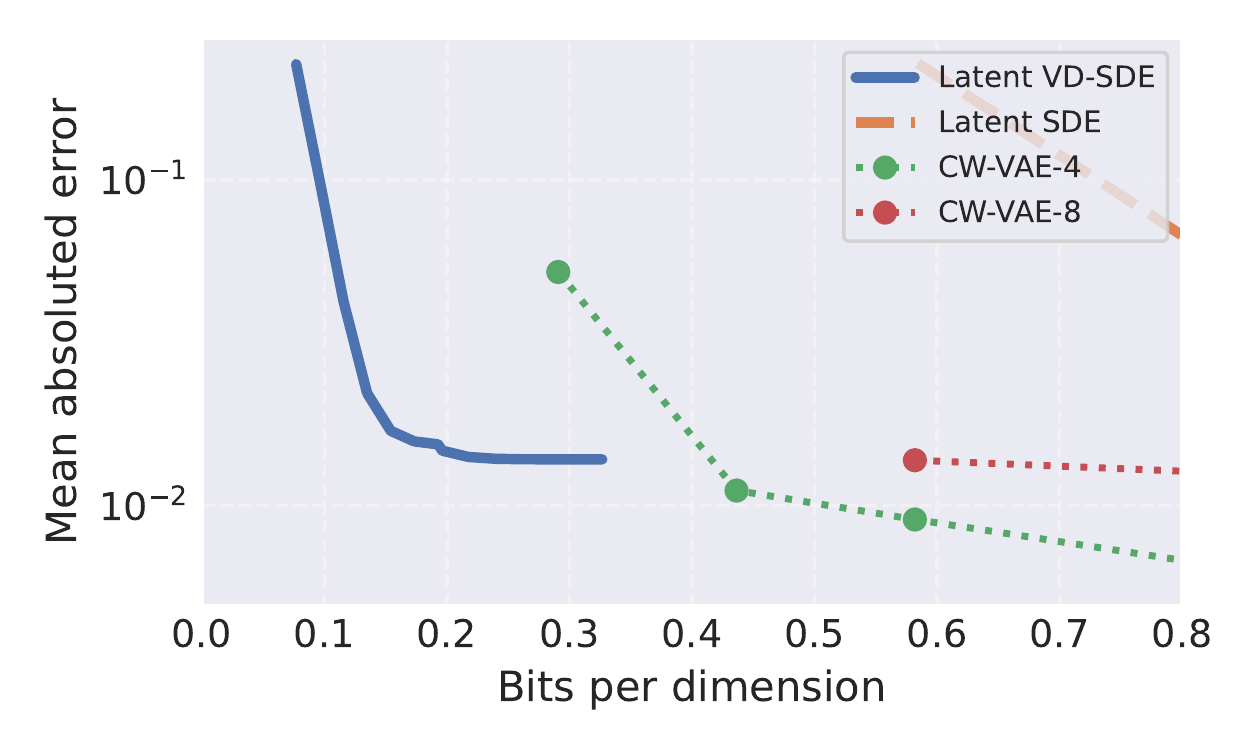}
         \caption*{AMASS}
    \end{subfigure}
    \vspace{-1em}
    \caption{Rate-distortion curves for motion capture sequence data sets. Even for very small CW-VAE models the low rate regime can not be reached. Whereas our model performs as well as the CW-VAE at rate=$0.3$ppd but only needs $0.15$bpd, a 2$\times$ compression boost. 
    }
    \label{fig:mocap_results}
\end{figure*}

\vspace{-0.7em}
\paragraph{Lossless Compression}

Table \ref{tab:lossless} contains results for lossless compression. We used a mixture of discretized logistics distribution~\citep{salimans2017pixelcnnpp} as the observation model. 
The estimated KL divergence is reported in bits while the total bits is reported in bits per pixel. 
As the majority of the bit rate is due to observation model, reducing the bit rate from the KL divergence in latent space does not significantly decrease compression cost. 
Nevertheless, it is worth noting that we can decrease KL significantly and that this may have more importance in future works, especially given the direction of using latent variables that are of much higher dimension than the observed data, \eg in the case of diffusion models.

\subsection{Motion Capture Compression}

\vspace{-0.7em}
\paragraph{Experiment details} 
Our experiments on motion capture data sets are similar, except we set the prior intensity value $\lambda$ between 10\% and 50\%, as there is less variability in the motion capture sequences so we can expect \textit{a priori} that the discretization can be coarser.
We use sequence lengths of $100$, set the observation standard deviation to $0.1$, and run with a learning rate of $0.0003$.

\Cref{fig:mocap_results} contains rate-distortion curves for the two motion capture data sets, where we report mean absolute error as a distortion metric.
Interestingly, not only does the discretization model learn to discretize at < 10\% of the original data sequence, the number of local latent dimensions is also a small percentage of the total. 
As such, the model learns to reduce the bit rate significantly compared to the original data format. 
This is expected as some of the motion capture sequences are smooth trajectories in space, though some erratic ones exist (see \Cref{fig:mocap_visual}).
When comparing to CW-VAE, we manually tuned the number of latent dimensions to be low in order to reduce the bit rate, which resulted in significant rate savings; however, this is a manual tuning step whereas our model automatically compressed its latent dimensions and discretization to arrive at the rates shown.

\section{Conclusion \& Discussion}

We propose Latent-VDSDE, a variationally-discretized Latent SDE, for neural compression. We infer a continuous-time representation for each data sequence, and infer a discretization grid that stores the local characteristics of the representation. This allows the model to automatically infer the number of sample points for each sequence.
Overall, we find that significant reduction in rates can be achieved from learning the discretization itself, and that our methods can significantly reduce the bit rate in the lossy compression regime without affecting the original distortion, and furthermore is competitive with a strong multi-level sequence model.
For smoother data sequences such as those of motion capture, we find that our model can achieve a very significant gain in efficiency from learning coarse discretizations. As an initial proof of concept, we find this to be a promising component for the neural sequence compression pipeline. 

\vspace{-1em}
\paragraph{Limitations \& Future Directions} In order to stabilize training when using \textsc{Reinforce} gradients, we had to resort to a two-stage training pipeline in order to have stable training, as discussed in Section \ref{sec:amortized_disc}. However, it may be possible to remove this constraint by directly using reparameterization gradients in continuous time \citep{chen2020learning}.
It would also be exciting to think about learning to discretize objects represented through higher dimensional functions, and to explore the usage of, or in combination with, domain-specific approaches and high-dimensional spatial quantization methods~\citep{ballard1999introduction,chen2022semi}.
As we extend these approaches to complex data sets such as LiDAR sequences and more, we need to handle the case where multiple sources of data can be continuously streamed, perhaps even at varying frequencies. A continuous-time approach should naturally handle such applications. Nevertheless, scaling up will require much stronger codec models, and perhaps combining with compression methods specific to such domains such as point cloud compression~\citep{quach2022survey} or video compression~\citep{agustsson2020scale,rippel2021elf}. 

\bibliographystyle{abbrvnat}
\bibliography{references}

\clearpage
\appendix
\onecolumn

\section{More background on data compression} \label{app:lossless}

The goal of lossless compression is to encode data points $x$ from a discrete space $\mathcal{X}$ as bit strings $m=\texttt{enc}(x) \in \{0,1\}^*$ of shortest possible length $\ell(m)$ such that $x$ can be decoded without loss of information. It can be shown, that complete prefix-free codes correspond to probability distributions $p_\theta(x)$ on $\mathcal{X}$, e.g.\ via a Shannon-Fano code \citep{shannon1948,fano1949,cover2012elements}. In this correspondence the lengths of the bit string of $x \in \mathcal{X}$ using the code $p_\theta(x)$ is (up to negligible rounding error) given by 
\begin{align}
    \ell(\texttt{enc}_{p_\theta}(x)) & = \lceil-\log p_\theta(x)\rceil.
\end{align} 

If we have several data points $x_1,\dots,x_n$ and we can assume that they come from a probabilistic source $x_i \sim p_d(x)$, $i=1,\dots,n$, then the smallest achievable average code length per point, in the large sample limit, using code $p_\theta(x)$, is given by the \emph{cross-entropy}:
\begin{align}
    H(p_d(x), p_\theta(x)) \triangleq \mathbf{E}_{x \sim p_d(x)} \left[ -\log p_\theta(x)\right].
\end{align}
This quantity is bounded from below (seen via Jensen's inequality) by the \emph{entropy} of the true probability distribution $p_d(x)$, i.e.\ when $p_\theta(x)=p_d(x)$ \citep{mackay2003information, cover2012elements}. Stream coders reach this bound up to a small constant $\epsilon$. A stream codec is a tuple of an inverse pair of functions, $\texttt{enc}_{p_\theta}$ and $\texttt{dec}_{p_\theta}$ that achieve near optimal compression rates on sequences of symbols.
\begin{equation}
\begin{aligned}
    &\texttt{enc}_{p_\theta}:m, x\mapsto m'\\
    &\texttt{dec}_{p_\theta}:m'\mapsto (m, x),
\end{aligned}
\end{equation}
such that $\ell(m')=\ell(m)+\log p_\theta(x) + \epsilon $.
Specifically in this paper, we will rely on \emph{Asymmetric numeral systems} (ANS) a type of stream coder that stores data in a stack-like data structure \citep{duda2009asymmetric}.  
Note that stream coders can also serve as a reservoir of randomness: Given any state $m$ we can decode from $m$ with a distribution $q$. This will return a random symbol $x \sim q$ and remove roughly $-\log q(x)$ bits of information from $m$.

\subsection{Bits-Back coding with ANS entropy coder} \label{app:lbitsback}

Bits-Back coding with ANS (BB-ANS) \citep{townsend2019practical} is a method for compressing sequences of symbols using latent variable models. Latent variable models are defined by their joint distribution $p(x, z)$, $z \in \latentsymbolspace$. 
Further, the joint distribution can be factorized as $p(z)p(x | z)$, and the probability mass functions (PMFs), cumulative distribution functions (CDFs), and inverse CDFs of $p(z)$ and $p(x | z)$ are tractable. However, computing the marginal distribution $p(x) = \sum_{z} p(z) p(x | z)$ is typically intractable.

To encode a symbol $x$ onto a message $m$ using BB-ANS, the sender first decodes a latent symbol $z$ from the message state using an approximate posterior distribution $q(z | x)$. The sender then pushes the pair $(x, z)$ onto the message using the distribution $p(x, z)$. The resulting message $m'$ has approximately $-\log p(x, z) + \log q(z | x)$ more bits than the original message $m$. Thus, the per-symbol rate saving of BB-ANS over a naive encoding method that directly encodes $(x, z)$ using $p(x, z)$ is $-\log q(z | x)$ bits. However, there is a one-time overhead for the initial message when encoding the first symbol.

The total bitrate of BB-ANS, defined as the number of bits in the final message per symbol encoded, will converge to the net bitrate, defined as the expected increase in message length per symbol, as the number of encoded symbols increases. The net bitrate of BB-ANS can be accurately predicted using the evidence lower bound (ELBO), 
\begin{equation}\label{eq:elbo}
\begin{aligned}
  \mathbb{E}_{(x,z) \sim p_d(x)q(z| x)}\left[-\log p(x, z) + \log q(z | x)\right]
  = H(p_d,p) + KL\left(q(z | x)||p(z | x)\right).
\end{aligned}
\end{equation}

More that BB-ANS has recently been extended to include hierarchical, and sequential latent variable models as well as achieve tighter bounds \citep{kingma2019bit,townsend2021lossless,ruan2021improving}. Aforementioned references are also excellent to dive into more details and subtleties of the method.

\subsection{Relative entropy coding}\label{app:relative_entropy_coding}

Relative entropy coding (REC) coding algorithms generate uniquely decodable codes $c$ representing a sample from $Q$ such that the codelength $|c|$ satisfies $\E[|c|] = O(\KL \left( Q \Vert P\right))$. It was shown that this is only possible if sender and receiver can rely on a sequence of random independent fair coin tosses $S = (s_1, s_2, . . .)$, refereed to as common randomness.
For practical purposes, $S$ stems from a public pseudo-random number generator.
\citet{li2018strong} were first in showing a general REC algorithm. For ordered arrival times $\{T_i\}$
of a homogeneous Poisson process on $\mathcal{R}$, and $X_i \sim P$, then
\begin{align}\label{eq:PFR}
    \argmin_{i\in \mathbf{N}} T_i \dfrac{dP}{dQ}(X_i)
\sim Q.
\end{align}
This leads to a coding scheme (PFR) in that a sample can be send by its index which minimizes \eqref{eq:PFR}. The resulting expected code length is bounded by $K \leq \E[|C|] \leq K + \log_2(K+1) + O(1)$, where $K = \KL \left( Q \Vert P\right)$.

Note that REC applies to both discrete and continuous distributions. Unfortunately, the runtime of PFR is so large that it renders the algorithm intractable in practice, specifically  the runtime requires $\Omega(2^{\KL \left( Q \Vert P\right)} )$ steps in expectation \citep{agustsson2020universally}.

Currently, multiple studies have explored computationally more efficient variants of this proposal that guarantees the same or approximately the same rate \citep{theis2021algorithms, flamich2022fast, flamich2020rec}.
In this study we applied fast REC (fREC) to the time variables \citep{flamich2022fast}. 
\citet{flamich2022fast} note that that \citet{li2018strong} formulation is equivalent to a search problem that can best be solved via the A-star sampling algorithm
\citep{maddison2014sampling}, thus develop a derivation of a variant Global Bound A-star sampling: the A-star coding algorithm.
\begin{align}
    \argmin_{i\in \mathbf{N}} T_i \dfrac{dP}{dQ}(X_i) 
    &= \argmax_{i\in \mathbf{N}} {- \log Ti + \log r(Xi)} \\
    &= \argmax_{i\in \mathbf{N}} 
{G_i + \log r(X_i)}
\end{align}

where $r = dQ/dP$, and $G_i$ is sampled according to $G_i \sim TG(0, G_{i-1})$ and $TG(\mu, \kappa)$ denotes the Gumbel distribution with mean
$\mu$ and unit scale, truncated to the interval $[-\infty, \kappa]$.
They show that for one-dimensional $Q$ and $P$ the expected code-length is optimal and the expected run-time is $O(D_\infty\left( Q \Vert P\right))$.

\section{Temporal Discretization Model Details} \label{app:q_tpp}

We construct a \text{SoftplusLogistic} distribution by transforming samples from a logistic distribution through a softplus function,
\begin{equation}
    \text{softplus}(z) = \log(1 + e^{z}).
\end{equation}
This is an invertible transformation, so we can exactly compute the probability density function and cumulative distribution functions. Specifically, let $g(x)$ be the inverse of a softplus function, \ie
\begin{equation}
    g(x) = \log(e^x - 1).
\end{equation}
Then we say $x \sim \text{SoftplusLogistic}(x \mid \mu, s)$ if $g(x) \sim \text{Logistic}(\mu , s)$. Furthermore, it has a probability density function
\begin{equation}
    p(x) = \tfrac{1}{\sigma(g(x))} p_\text{logistic}(g(x) \mid \mu, s)
\end{equation}
where $p_\text{logistic}$ is the probability density function of a logistic distribution with mean $\mu$ and scale $s$, and $\sigma$ is the sigmoid function. It also has the cumulative distribution function (CDF)
\begin{equation}
    F(x) = \sigma\left(\tfrac{g(x) - \mu}{s}\right)
\end{equation}
Having access to the CDF allows us to truncate the distribution at any value and restrict the domain to be $(0, t_\text{max}]$. This results in the modified density function
\begin{equation}
    p(x \mid x \leq t_\text{max}) = \tfrac{p(x)}{F(x)} \mathbbm{1}_{x \leq t_\text{max}}
\end{equation}
We set $t_\text{max} = 1.0$ in all of our experiments, in order to control the maximum possible time interval between grid points.

\section{Extra Figures}\label{app:extra_figs}

\begin{figure}
    \centering
    \raisebox{2em}{\rotatebox[origin=t]{90}{\small CMU MoCap}}%
    \includegraphics[width=0.47\linewidth]{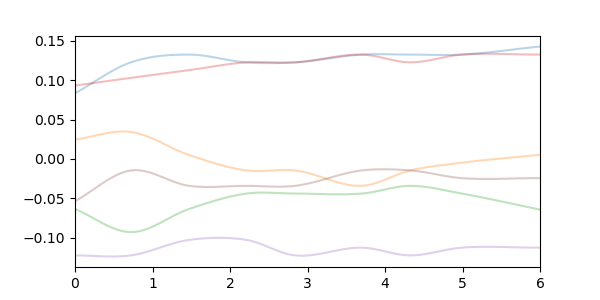}
    \includegraphics[width=0.47\linewidth]{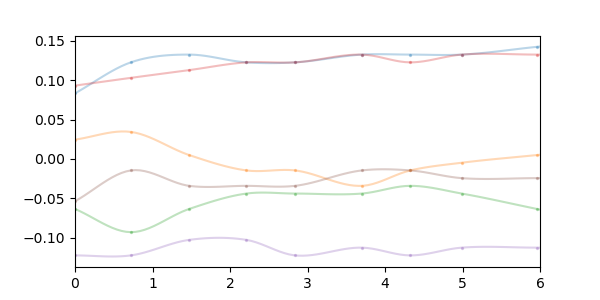}\\
    \raisebox{3em}{\rotatebox[origin=t]{90}{\small Moving MNIST}}%
    \begin{subfigure}[b]{0.47\linewidth}
         \centering
         \includegraphics[width=\linewidth]{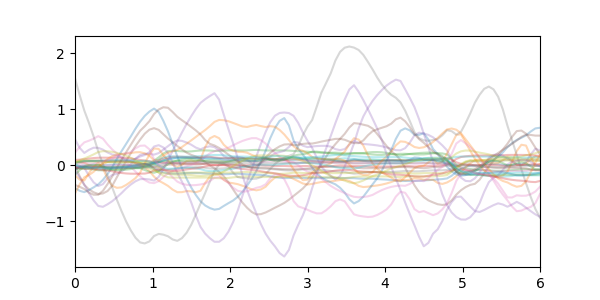}
         \caption{Full sample path}
    \end{subfigure}
    \begin{subfigure}[b]{0.47\linewidth}
         \centering
         \includegraphics[width=\linewidth]{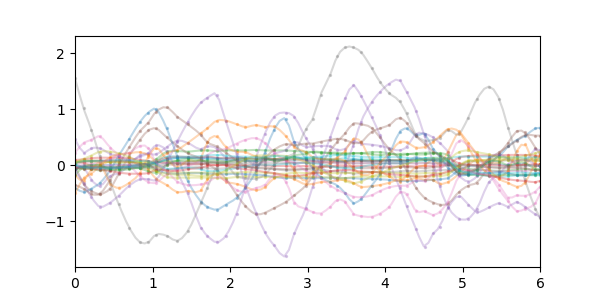}
         \caption{Learned Discretization}
    \end{subfigure}
    \caption{\textbf{Learned discretization}. Sequence complexity determines the sparsity of the discretization grid. A single sample for each latent dimension is plotted where dots indicate discretized points. Major curves are captured while minuscule changes are smoothed out through the cubic spline interpolation.}
\end{figure}

\begin{figure}
    \centering
    \begin{subfigure}[b]{0.48\linewidth}
    \centering
    \includegraphics[width=\linewidth]{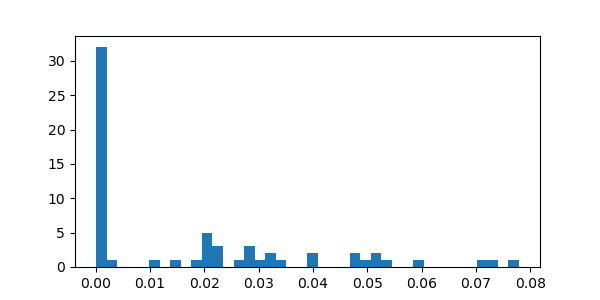}
    \caption*{$\nu$}
    \end{subfigure}
    \begin{subfigure}[b]{0.48\linewidth}
    \centering
    \includegraphics[width=\linewidth]{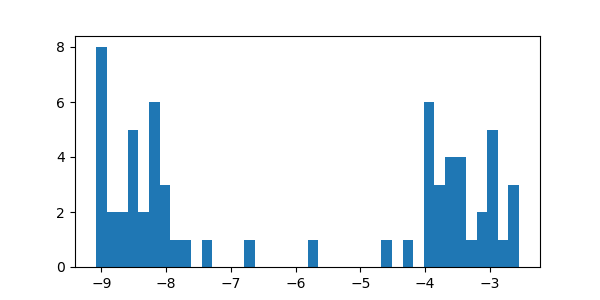}
    \caption*{$\log \nu$}
    \end{subfigure}
    \caption{\textbf{Global vs Local Representations.} Histogram of values of $\nu$ for each latent dimension. We found that most trained models automatically end up with two clusters of $\nu$ values, with one cluster containing values that are effectively zero. See \cref{fig:nu_clusters}.}
    \label{fig:nu_clusters}
\end{figure}

\begin{figure}
    \centering
    \begin{subfigure}[b]{0.49\linewidth}
         \centering
         \includegraphics[width=\linewidth]{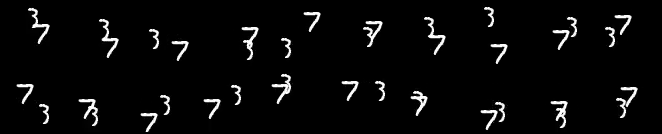}
    \end{subfigure}
    \begin{subfigure}[b]{0.49\linewidth}
         \centering
         \includegraphics[width=\linewidth]{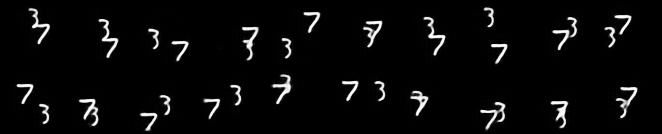}
    \end{subfigure}\\
    \begin{subfigure}[b]{0.49\linewidth}
         \centering
         \includegraphics[width=\linewidth]{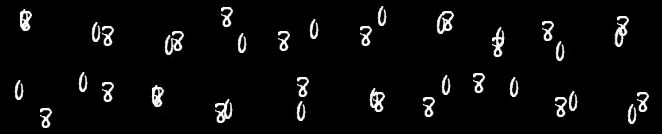}
    \end{subfigure}
    \begin{subfigure}[b]{0.49\linewidth}
         \centering
         \includegraphics[width=\linewidth]{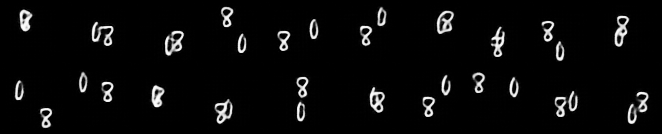}
    \end{subfigure}\\
    \begin{subfigure}[b]{0.49\linewidth}
         \centering
         \includegraphics[width=\linewidth]{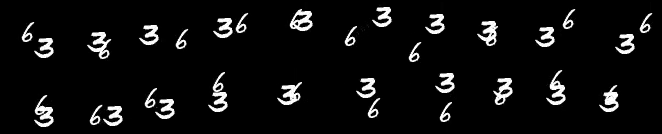}
         \caption*{Original}
    \end{subfigure}
    \begin{subfigure}[b]{0.49\linewidth}
         \centering
         \includegraphics[width=\linewidth]{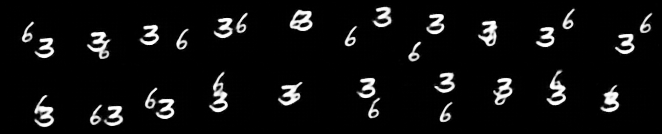}
         \caption*{Compressed}
    \end{subfigure}\\
    \caption{Original sequences and their reconstruction for Moving MNIST.}
    \label{fig:reconstructions_mmnist}
\end{figure}

\begin{figure}
    \centering
    \begin{subfigure}[b]{0.49\linewidth}
         \centering
         \includegraphics[width=\linewidth]{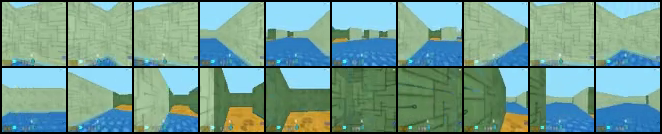}
    \end{subfigure}
    \begin{subfigure}[b]{0.49\linewidth}
         \centering
         \includegraphics[width=\linewidth]{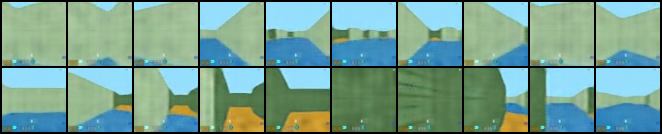}
    \end{subfigure}\\
    \begin{subfigure}[b]{0.49\linewidth}
         \centering
         \includegraphics[width=\linewidth]{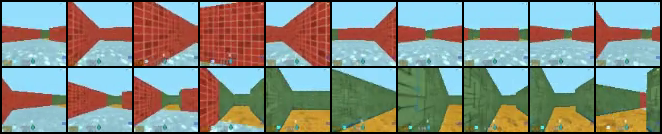}
    \end{subfigure}
    \begin{subfigure}[b]{0.49\linewidth}
         \centering
         \includegraphics[width=\linewidth]{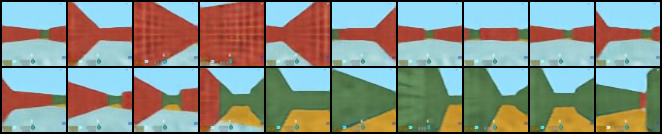}
    \end{subfigure}\\
    \begin{subfigure}[b]{0.49\linewidth}
         \centering
         \includegraphics[width=\linewidth]{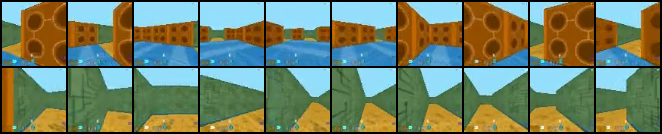}
         \caption*{Original}
    \end{subfigure}
    \begin{subfigure}[b]{0.49\linewidth}
         \centering
         \includegraphics[width=\linewidth]{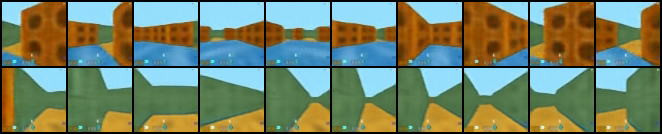}
         \caption*{Compressed}
    \end{subfigure}\\
    \caption{Original sequences and their reconstruction for GQN Mazes.}
    \label{fig:reconstructions_mazes}
\end{figure}

\begin{figure}
    \centering
    \begin{subfigure}[b]{0.49\linewidth}
         \centering
         \includegraphics[width=\linewidth]{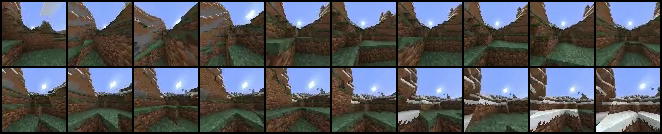}
    \end{subfigure}
    \begin{subfigure}[b]{0.49\linewidth}
         \centering
         \includegraphics[width=\linewidth]{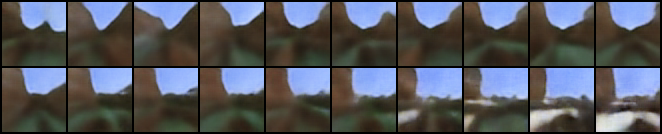}
    \end{subfigure}\\
    \begin{subfigure}[b]{0.49\linewidth}
         \centering
         \includegraphics[width=\linewidth]{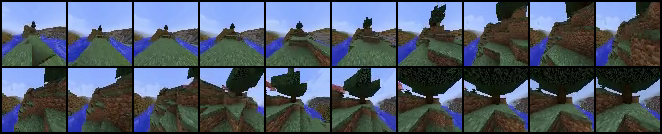}
    \end{subfigure}
    \begin{subfigure}[b]{0.49\linewidth}
         \centering
         \includegraphics[width=\linewidth]{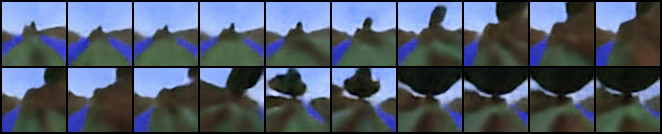}
    \end{subfigure}\\
    \begin{subfigure}[b]{0.49\linewidth}
         \centering
         \includegraphics[width=\linewidth]{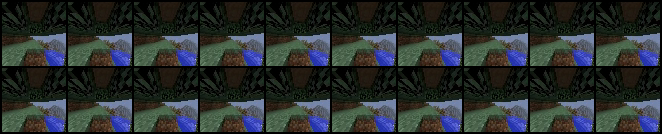}
         \caption*{Original}
    \end{subfigure}
    \begin{subfigure}[b]{0.49\linewidth}
         \centering
         \includegraphics[width=\linewidth]{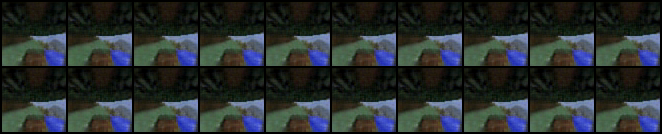}
         \caption*{Compressed}
    \end{subfigure}\\
    \caption{Original sequences and their reconstruction for MineRL.}
    \label{fig:reconstructions_minerl}
\end{figure}

\end{document}